\pdfoutput=1

\documentclass[11pt]{article}

\usepackage{ACL2023}

\usepackage{times}
\usepackage{latexsym}

\usepackage[T1]{fontenc}

\usepackage[utf8]{inputenc}

\usepackage{microtype}

\usepackage{inconsolata}

\usepackage{url}

\usepackage{graphicx}
\usepackage{amsfonts}
\usepackage{amsmath}
\usepackage{amssymb}
\usepackage{pifont}
\usepackage{nicefrac}
\usepackage{multirow}
\usepackage{subcaption}
\usepackage[font=small]{caption}
\usepackage{soul}
\usepackage{booktabs}
\usepackage{xcolor}
\usepackage{comment}
\usepackage{sidecap}
\usepackage{xspace}
\usepackage[most]{tcolorbox}
\usepackage{enumitem}
\usepackage{ulem}
\usepackage{makecell}
\usepackage[textsize=tiny,textwidth=24mm]{todonotes}

\usepackage[utf8]{inputenc}
\usepackage[T1]{fontenc}
\usepackage{CJKutf8}

\newcommand{\chinese}[1]{{\begin{CJK*}{UTF8}{gkai} #1 \end{CJK*}}}

\definecolor{bblue}{HTML}{4F81BD}
\definecolor{rred}{HTML}{c4260b}
\definecolor{ggreen}{HTML}{098c1f}
\definecolor{ppurple}{HTML}{9F4C7C}
\definecolor{oorange}{HTML}{F79646}

\DeclareRobustCommand{\hlred}[1]{{\textcolor{rred}{#1}}}
\DeclareRobustCommand{\hlblue}[1]{{\textcolor{bblue}{#1}}}

\newcommand{\hidden}[1]{\vphantom{#1}}
\newcommand{\nosig}[1]{\colorbox{gray!30}{#1}}
\newcommand{\lessthree}[1]{{#1}}
\newcommand{\geaterfive}[1]{\colorbox{red!30}{#1}}
\newcommand{\posres}[1]{\colorbox{bblue!30}{#1}}

\newcommand{\glm}{{GLM-130B}\xspace}
\newcommand{\flores}{{FLORES}\xspace}
\newcommand{\llms}{{LLMs}\xspace}
\newcommand{\llm}{{LLM}\xspace}

\usepackage{tablefootnote}

\title{Prompting Large Language Model for Machine Translation: A Case Study}

\author{Biao Zhang \quad Barry Haddow \quad Alexandra Birch \bigskip\\
  School of Informatics, University of Edinburgh \\
  {\texttt{b.zhang@ed.ac.uk, bhaddow@inf.ed.ac.uk, a.birch@ed.ac.uk}}
  }

\begin{document}
\maketitle
\begin{abstract}

Research on prompting has shown excellent performance with little or even no supervised training across many tasks. However, prompting for machine translation is still under-explored in the literature. We fill this gap by offering a systematic study on prompting strategies for translation, examining various factors for prompt template and demonstration example selection. We further explore the use of monolingual data and the feasibility of cross-lingual, cross-domain, and sentence-to-document transfer learning in prompting. Extensive experiments with \glm~\cite{zeng2022glm130b} as the testbed show that 1) the number and the quality of prompt examples matter, where using suboptimal examples degenerates translation;  2) several features of prompt examples, such as semantic similarity, show significant Spearman correlation with their prompting performance; yet, none of the correlations are strong enough; 3) using pseudo parallel prompt examples constructed from monolingual data via zero-shot prompting could improve translation; and 4) improved performance is achievable by transferring knowledge from prompt examples selected in other settings. We finally provide an analysis on the model outputs and discuss several problems that prompting still suffers from.


\end{abstract}

\section{Introduction}

Large language models (\llms) pretrained on massive unlabeled corpora have shown impressive emergent abilities under model scaling which enable prompting for downstream applications~\cite{gpt3,kaplan2020scaling,wei2022emergent,zhang2022opt,chowdhery2022palm}. Different from task-specific finetuning, prompting constructs task-specific prompts by rephrasing test examples with descriptive task instructions and executes the task by feeding prompts to \llms directly. It can be further enhanced through in-context learning by providing a few labeled examples (or prompt examples) as 
a demonstration~\cite{gpt3}. As a new paradigm, prompting \llms has achieved state-of-the-art performance over a range of natural language processing (NLP) tasks~\cite{chung2022scaling,goyal2022news,wei2022chain,chowdhery2022palm}.

In this paper, we focus on prompting \llms for machine translation (MT). MT represents a complex task requiring transforming a source input into its semantically equivalent target output in a different language, which combines sequence understanding and generation. It offers a unique platform to assess the cross-lingual generation capability of \llms, and the assessment may shed light on pretraining/finetuning algorithm design for achieving universal \llms~\cite{chowdhery2022palm}. While a few studies have reported translation results~\cite{gpt3,chowdhery2022palm}, a systematic study on how prompting works for MT is still missing in the literature.

We aim at filling this gap by thoroughly examining different prompting setups using the recently released \glm~\cite{zeng2022glm130b}, particularly concerning three aspects: \textit{the prompting strategy}, \textit{the use of unlabeled/monolingual data}, and \textit{the feasibility of transfer learning}. Prompting has shown varying sensitivity to the choice of prompt templates and examples~\cite{pmlr-v139-zhao21c}. For MT, prior studies adopted different templates~\cite{gpt3,wei2022finetuned,chowdhery2022palm}, and we reevaluate them to figure out the optimal one. We further design a set of features for prompt examples and explore which one(s) could explain the prompting performance, according to which we develop the example selection strategy.

Since leveraging monolingual data to improve MT has long been of interest, we would like to determine whether and how such data can be used in prompt example construction. We make a step in this direction by studying the effect of data augmentation using back-/forward-translation~\cite{sennrich-etal-2016-improving,zhang2016exploiting} via zero-shot prompting. In addition, neural MT and pretrained \llms have shown encouraging transfer abilities~\cite{devlin-etal-2019-bert,arivazhagan2019massively,zhang-etal-2020-improving,xue2021mt5} but transfer learning for prompting has received little attention. Whether prompt examples are transferable across different settings, such as from one domain/language pair to another and from sentence-level examples to document-level translation, is yet to be addressed.

We address the above concerns with \glm as the testbed and conduct extensive experiments on \flores and WMT evaluation sets. We mainly study translation for three languages: English, German and Chinese. We also provide a quantitative and qualitative analysis to disclose problems when prompting for MT, which might offer insights for future study. Our main findings are listed as below:
\begin{itemize}
    \item Prompting performance varies greatly across templates, and language-specific templates mainly work when translating into languages \llms are pretrained on. An English template in a simple form works best for MT.
    \item Several features of prompt examples, such as sequence length, language model score, and semantic similarity, correlate significantly with its prompting performance while the correlation strength is weak in general. Selecting examples based on these features can outperform the random strategy, but not consistently.
    \item Using monolingual examples for prompting hurts translation. By contrast, constructing pseudo parallel examples via back-/forward-translation is a good option. Back-translation performs better and is more robust.
    \item Prompting shows some degree of transferability. Using demonstrations from other settings can improve translation over the zero-shot counterpart, while the superiority of a demonstration in one setting can hardly generalize to another.
    \item Prompting for MT still suffers from copying, mistranslation of entities, hallucination, inferior direct non-English translation, and prompt trap where translating the prompt itself via prompting becomes non-trivial.
\end{itemize}

\section{Setup}

\paragraph{Prompting for MT} Given a pretrained and \textit{fixed} \llm $\mathcal{L}$, MT prompting first converts each test input $X$ to a prompt according to a template $\mathcal{T}$ and then generate the translation $Y$ by feeding the prompt to $\mathcal{L}$. In this study, we consider \textit{zero-shot} and \textit{few-shot} prompting for translation. 

Zero-shot prompting only has access to the test input $X$, while few-shot prompting assumes that a few extra labeled examples (or \textit{prompt/demonstration examples}) $\mathcal{D}^P = \{X_i^\prime, Y_i^\prime\}_{i=1}^K$ are available and can be used as a \textit{demonstration}. Particularly, we adopt the following template for zero-shot prompting based on the results in Section \ref{sec:prompting_strategy}:
\begin{equation}
    \texttt{[src]: $X$ [tgt]: }
\end{equation}
where \texttt{[src]} and \texttt{[tgt]} denote \textit{test language(s)}, i.e., the source and target language name of the test language pair, respectively. For few-show prompting, we concatenate the given prompt examples:
\begin{align}
    &\texttt{[psrc]: $X_1^\prime$ [ptgt]: $Y_1^\prime \ldots$ [psrc]: $X_K^\prime$} \nonumber \\
    & \texttt{[ptgt]: $Y_K^\prime$ [src]: $X$ [tgt]: } \label{eq:few_shot}
\end{align}
where \texttt{[psrc]} and \texttt{[ptgt]} denote \textit{prompt language(s)}, i.e., the source and target language name of the prompt example, respectively. By default, prompt examples and test data are in the same language pair. However, when considering cross-lingual transfer for prompting, prompt examples might be in a different language pair.

We also explore template language, which denotes the language in which the template is expressed. For example, the Chinese template ``\chinese{中文：$X$ 英文：}'' represents the Chinese counterpart of the following English template ``\textit{Chinese: $X$ English: }''.

\begin{table*}[t]
    \centering
    \small
    \begin{tabular}{llrrrrrr}
    \toprule
    \multirow{2}{*}{ID} & \multirow{2}{*}{Template (in English)} & \multicolumn{2}{c}{English} & \multicolumn{2}{c}{German} & \multicolumn{2}{c}{Chinese} \\
    \cmidrule(lr){3-4} \cmidrule(lr){5-6} \cmidrule(lr){7-8}
    & & \multicolumn{1}{c}{w/o} & \multicolumn{1}{c}{w/} & \multicolumn{1}{c}{w/o} & \multicolumn{1}{c}{w/} & \multicolumn{1}{c}{w/o} & \multicolumn{1}{c}{w/} \\
    \midrule
    A & \texttt{[src]: [input] $\diamond$ [tgt]:} & \textbf{38.78} & \textbf{31.17} & -26.15 & -16.48 & \textbf{14.82} & \textbf{-1.08} \\
    B & \texttt{[input] $\diamond$ [tgt]:} & -88.62 & -85.35 & -135.97 & -99.65 & -66.55 &  -85.84 \\
    C & \texttt{[input] $\diamond$} {Translate to} \texttt{[tgt]: } & -87.63 & -68.75 & -106.30 & -73.23 & -63.38 & -70.91 \\
    D & \texttt{[input] $\diamond$} {Translate from} \texttt{[src]} {to} \texttt{[tgt]: } & -113.80 & -89.16 & -153.80 & -130.65 & -76.79 & -67.71 \\
    E & \texttt{[src]: [input] $\diamond$} {Translate to} \texttt{[tgt]: } & 20.81 & 16.69 & \textbf{-24.33} & \textbf{-5.68} & -8.61 & -30.38 \\
    F & \texttt{[src]: [input] $\diamond$} {Translate from} \texttt{[src]} {to} \texttt{[tgt]: } & -27.14 & -6.88 & -34.36 & -9.22 & -32.22 & -44.95 \\
    \bottomrule
    \end{tabular}
    \caption{\label{tab:res_template} COMET scores averaged over 6 language pairs for \textit{zero-shot} prompting with different templates and different template languages on Wiki Ablation sets. \textit{w/} and \textit{w/o} denote whether adding line breaks into the template or not; $\diamond$ indicates the position of the line break. \texttt{[src]} and \texttt{[tgt]} denote source and target test language name, respectively, and \texttt{[input]} denotes the test input; all of them are placeholders. \textit{English, German} and \textit{Chinese} indicate template languages. Best results are shown in \textbf{bold}.}
\end{table*}

\begin{figure*}[t]
  \centering
  \small
  \resizebox{\textwidth}{!}
  {\includegraphics[scale=0.50]{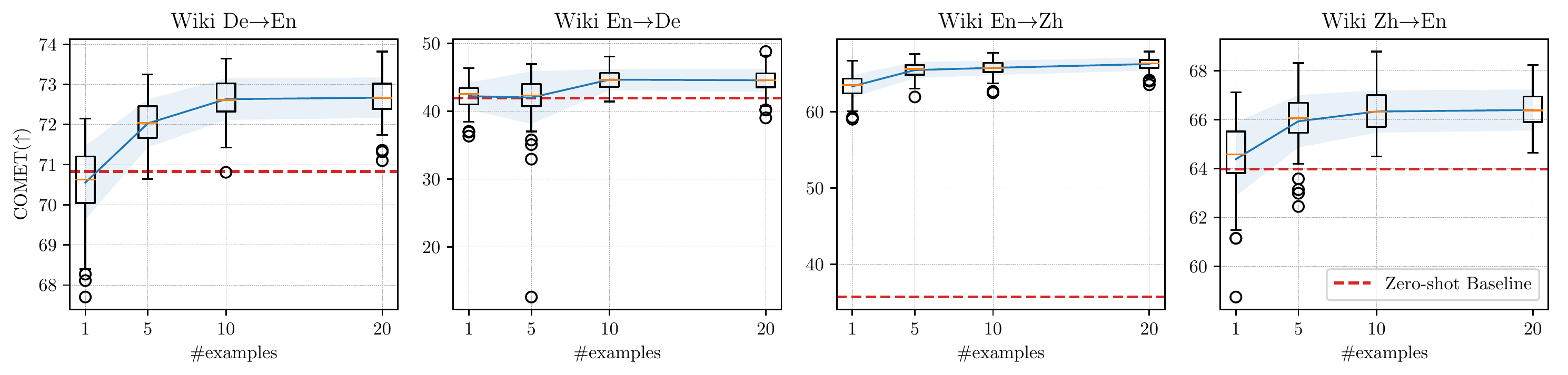}}
  \caption{\label{fig:num_examples} COMET scores for \textit{few-shot} prompting as a function of the number of prompt examples ($K=1,5,10,20$) on Wiki Ablation sets. For each setup, we randomly sample 100 times from the example pool and show the performance distribution via box plots. Dashed red line denotes the zero-shot baseline; blue curve and shadow area denote the mean and standard deviation.}
\end{figure*}

\paragraph{Setting} We experiment with \glm, a \llm with 130B parameters pretrained on Chinese and English monolingual corpora, which was reported to outperform GPT-3 and OPT-175B on several NLP tasks~\cite{zeng2022glm130b}. Note \glm is a raw \llm without any further finetuning. We use its INT4-quantized version, which is more affordable and suffers little performance degradation. We adopt beam search for MT with a beam size of 2, and perform experiments with 4 RTX 3090 and A100-40G GPUs.

We work on three languages: English (En), German (De), and Chinese (Zh). We perform major analysis on \flores~\citep[Wiki domain, En-De-Zh, ][]{nllb2022} and WMT21~\citep[News domain, En-De, En-Zh, ][]{akhbardeh-EtAl:2021:WMT}, and also report results on Multi-Domain~\citep[IT, Law and Medical domain, De-En, ][]{aharoni-goldberg-2020-unsupervised} to examine domain robustness and transfer ability, and PDC~\citep[News domain, Zh$\rightarrow$En, ][]{sun-etal-2022-rethinking} for document-level translation. To understand the relation between prompt examples and their prompting performance, we construct an \textbf{Ablation} set for Wiki, WMT and Multi-Domain (IT and Medical) based on the dev set of \flores, WMT21 and Multi-Domain, separately, where we randomly sample 100 instances as the ablation test set and use the rest as the default example selection pool. To distinguish, we will refer to the official dev and test set as \textbf{Full} set. Detailed statistics are listed in Table \ref{tab:data_statistics}, Appendix.

We evaluate translation performance using both a surface-based metric, detokenized BLEU$\uparrow$ from SacreBLEU~\cite{post-2018-call}, and a model-based metric, COMET$\uparrow$ from \texttt{unbabel-comet} with \textit{wmt20-comet-da}~\cite{rei-etal-2020-comet}. 

\section{Prompting Strategy for MT}\label{sec:prompting_strategy}

To perform MT, prompting needs to cast the translation problem into a language modeling problem via the prompt. Thus, the format of the prompt, including its wording, directly affects how \llm understands the task and its behavior. For MT, we are interested in the following research questions:
\begin{itemize}
    \item Which template should we use for MT prompting? And what language for the template?
    \item Does demonstration matter for MT prompting? How to select optimal prompt examples?
\end{itemize}
We address them through extensive experiments on Wiki Ablation sets.

\paragraph{Zero-shot prompting performance varies greatly across templates.} We start with zero-shot prompting and explore the effect of different templates. Depending on how to describe MT and partially inspired by prior studies~\cite{gpt3,chowdhery2022palm,wei2022finetuned}, we compare 6 templates and evaluate them on the Wiki Ablation sets covering 6 language pairs (En$\leftrightarrow$De, En$\leftrightarrow$Zh, De$\leftrightarrow$Zh). Table \ref{tab:res_template} shows the results (we list detailed results in Table \ref{tab:res_template_detailed_comet}, Appendix). The template affects zero-shot quality substantially, and the simple template \textcircled{A} in English specifying just the source and target language name achieves the best overall results. In follow-up experiments, we thus focus on template \textcircled{A}.

\begin{table}[t]
    \setlength{\tabcolsep}{9pt}
       \small
       \centering
        \begin{tabular}{lrrrr}
        \toprule
        \multirow{4}{*}{Feature} & \multicolumn{2}{c}{BLEU} & \multicolumn{2}{c}{COMET} \\
        \cmidrule(lr){2-3} \cmidrule(lr){4-5}
        & \multicolumn{1}{c}{HQ} & \multicolumn{1}{c}{ + LQ} & HQ & + LQ \\
        \midrule
        SLength & 0.21 & 0.31 & 0.14  & 0.26 \\
        TLength & \textbf{0.23} & 0.32 & \textbf{0.17}  &  0.29 \\
        LMScore & 0.20 & \textbf{0.33} & 0.14 &  \textbf{0.31} \\
        MTScore & 0.04 & 0.14 &  0.11 &  0.19 \\
        SemScore & 0.19 & 0.30 & 0.16  &  0.30 \\
        CaseSemScore-Src & 0.14 & 0.29 &  0.11 &  0.28 \\
        CaseSemScore-Tgt & 0.14 & 0.30 &  0.14 &  \textbf{0.31} \\
        \bottomrule
        \end{tabular}
        \caption{\label{tab:1_shot_factor_correlation} Spearman's $\rho$ between demonstration features and their prompting performance for \textit{1-shot} prompting on Wiki Ablation sets. We randomly sample 600 demonstrations from each pool to calculate the correlation. \textit{HQ}: examples are from the default high-quality pool; \textit{LQ}: examples are from the low-quality pool based on WikiMatrix.v1.}
 \end{table}
 
\begin{figure}[t]
   \centering
   \resizebox{\columnwidth}{!}
    {\includegraphics[scale=0.47]{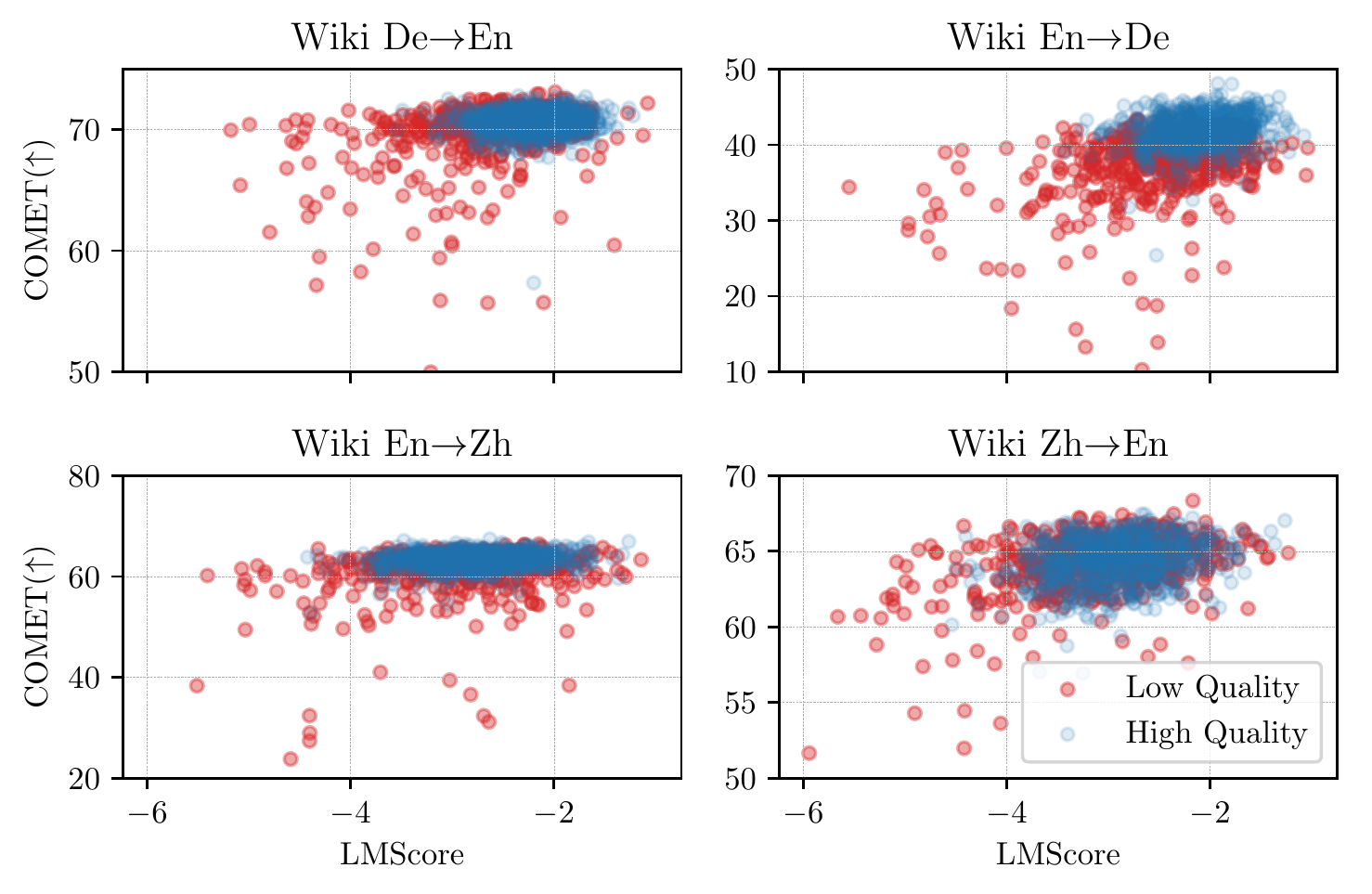}}
    \caption{\label{fig:1_shot_factor_correlation} Visualization between COMET and LMScore for \textit{1-shot} prompting on Wiki Ablation sets. While correlations are significant, data points are scattered like clouds.} 
\end{figure}

\paragraph{Language-specific template delivers mixed results.} Table \ref{tab:res_template} also shows the prompting results of German and Chinese templates, which often largely underperform their English counterparts. Since German is not a major pretraining language in \glm, a German template degenerates the translation substantially. By contrast, a Chinese template yields improved performance when translating into Chinese (see Table \ref{tab:res_template_detailed_comet}). Still, an English template works best on average. 

The preference of \glm to English template also shows that the level of language understanding and cross-lingual ability in \glm varies across languages, even though it's pretrained on the same amount of monolingual Chinese and English tokens. This might be caused by the fact that English is used more globally than Chinese, but might also suggest that improving the language understanding of \llm requires more advanced training algorithms beyond scaling training data.

\begin{figure}[t]
  \centering
  \small
  {\includegraphics[scale=0.50]{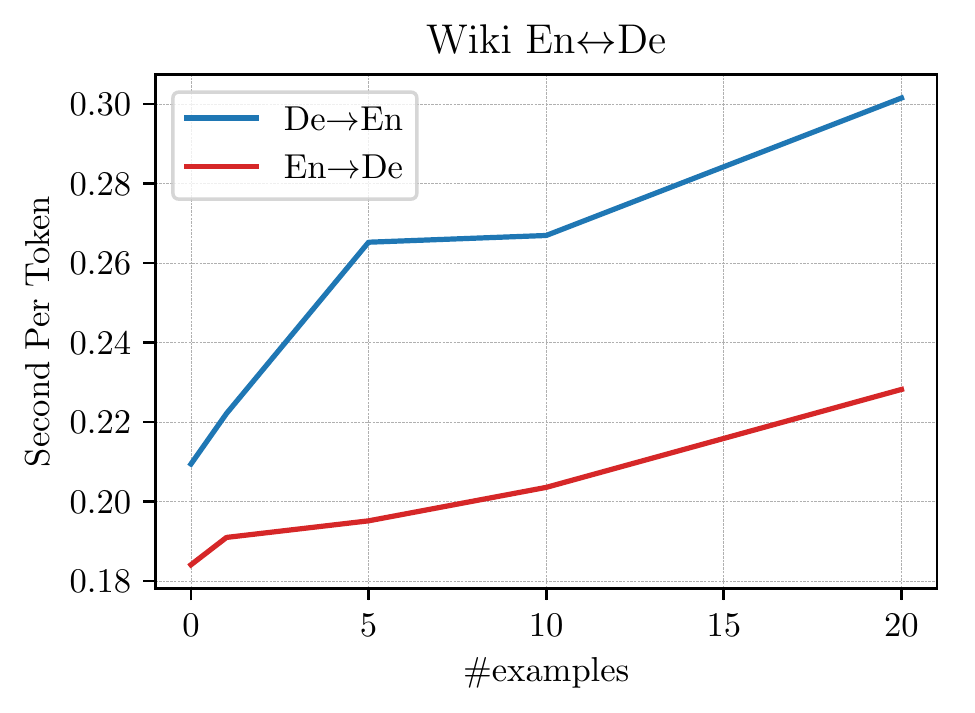}}
  \caption{\label{fig:timing} Inference time per token in seconds for \textit{zero-/few-shot} prompting on Wiki En-De Ablation sets. Numbers are averaged over 3 runs with 3 distinct demonstrations on 4 A100-40G GPUs.}
\end{figure}

\paragraph{Using more prompt examples for demonstration improves translation significantly on average.} We next study few-shot prompting following the template \textcircled{A} but in format (\ref{eq:few_shot}) with $K$ varying from 1 to 20. We evaluate multiple demonstrations for each $K$ via random sampling to reduce data biases. Figure \ref{fig:num_examples} shows that the more examples used, the better average performance (more results are shown in Figure \ref{fig:num_examples_extra}, Appendix), albeit at the cost of using more GPU memory and increasing the inference time per token as in Figure \ref{fig:timing}.

\paragraph{The performance of demonstration is not stable.} However, we also see high performance variance under the same $K$. It's possible that a demonstration with 5 examples outperforms its 10 or 20 counterpart. Figure \ref{fig:num_examples} also shows that 1-shot prompting underperforms zero-shot prompting in many cases, even on average. This echoes with previous findings on other NLP tasks~\cite{pmlr-v139-zhao21c,liu-etal-2022-makes} and also highlights the significance of developing effective example selection strategies.

Note that few-shot prompting greatly improves translation into Chinese. The reason based on our manual analysis is that the zero-shot baseline tends to translate into traditional Chinese with messy codes, where prompt examples help (the reference text is always simplified Chinese).


\paragraph{Several features correlate with prompting performance significantly yet weakly.} We thus turn to explore example selection for prompting. Our idea is to extract a couple of diverse features from demonstration and examine whether any of them are informative enough to be used as an indicator for the selection. In this study, we simplify our analysis by focusing on 1-shot prompting, which ignores the ordering of prompt examples (we leave few-shot analysis to future). Particularly, we extract and analyze 7 features of a demonstration: 
\begin{description}
    \item[S(T)Length] the number of source (target) tokens;
    \item[LMScore] \glm-based, length-normalized log likelihood of the demonstration;
    \item[MTScore] translation quality of the prompt example from COMET QE model \textit{wmt20-comet-qe-da}~\cite{rei-etal-2020-comet};
    \item[SemScore] semantic score based on the cosine similarity of the demonstration's source and target sentence embeddings from LASER2~\cite{heffernan2022bitext};
    \item[CaseSemScore-Src] similarity to the input that averages over SemScores between the test input and the demonstration's source;
    \item[CaseSemScore-Tgt] similar to CaseSemScore-Src but compares to demonstration's target;
\end{description}
We sample multiple demonstrations randomly and inspect the Spearman's correlation between feature values and prompting performance. We consider high-quality and low-quality pool for sampling.

\begin{table}[t]
    \centering
    \small
    \begin{tabular}{lrrrr}
    \toprule
    \multirow{2}{*}{Method} & \multicolumn{2}{c}{Wiki} & \multicolumn{2}{c}{WMT} \\
    \cmidrule(lr){2-3} \cmidrule(lr){4-5} 
    & BLEU & COMET & BLEU & COMET \\
    \midrule
    Zero-Shot & 24.08 & 33.92  & 20.38 & 17.97 \\
    \midrule
    \multicolumn{5}{l}{\it 1-Shot Translation (high-quality pool)}  \vspace{0.1cm} \\
    Random & 26.31\hidden{0.36} & 48.29\hidden{0.99} & 21.27\hidden{0.53} & 30.70\hidden{2.72} \\ 
    SemScore & \underline{26.73}\hidden{0.23} & \underline{49.34}\hidden{0.79} & \underline{21.82}\hidden{0.54} & \underline{31.28}\hidden{1.78} \\ 
    LMScore & 26.48\hidden{0.44} & 47.92\hidden{0.92} & 21.59\hidden{0.10} & 30.81\hidden{0.34} \\
    TLength & 26.54\hidden{0.38} & 48.73\hidden{1.02} & 21.29\hidden{0.38} & 30.68\hidden{1.44} \\
    \midrule
    \multicolumn{5}{l}{\it 5-Shot Translation (high-quality pool)}  \vspace{0.1cm} \\
    Random & \textbf{27.46}\hidden{0.25} & 51.11\hidden{0.57} & 21.82\hidden{0.48} & 33.87\hidden{1.26} \\
    SemScore & 27.36\hidden{0.20} & \textbf{51.66}\hidden{0.32}  & \textbf{22.37}\hidden{0.24} & 34.30\hidden{0.72} \\
    LMScore & 27.17\hidden{0.25} & 50.65\hidden{0.40} & 22.04\hidden{0.25} & \textbf{35.19}\hidden{0.53} \\
    TLength & 27.08\hidden{0.24} & 50.50\hidden{0.28} & 21.75\hidden{0.08} & 34.29\hidden{0.56} \\ 
    \midrule
    \multicolumn{5}{l}{\it 1-shot Translation (Low-quality Pool)}  \vspace{0.1cm} \\
    Random & 24.75\hidden{1.79} & 38.86\hidden{9.48}  & 22.06\hidden{0.86} & 30.70\hidden{2.25} \\ 
    Ours & \underline{24.94}\hidden{1.41} & \underline{39.88}\hidden{8.18} & \underline{22.23}\hidden{0.42} & \underline{30.87}\hidden{1.62} \\ 
    \bottomrule
    \end{tabular}
    \caption{\label{tab:res_wiki_wmt} BLEU and COMET scores for \textit{zero-shot and few-shot} prompting on Wiki and WMT Full sets with different selection strategies. \textit{Ours}: the proposed combined strategy; \textit{Random}: random sampling; \textit{SemScore, LMScore} and \textit{TLength} denote selecting top-ranked examples based on the corresponding feature values. We select 3 demonstrations for each translation direction and report average performance; the final score is further averaged over different language pairs. \underline{Underlined} results denote the best in each section, while \textbf{Bold} results are the overall best.}
\end{table}

Table \ref{tab:1_shot_factor_correlation} summarizes the results and Figure \ref{fig:1_shot_factor_correlation} illustrates the relation between COMET and LMScore (more results are given in Table \ref{tab:1_shot_factor_correlation_bleu} and Figures \ref{fig:1_shot_factor_correlation_bleu}, \ref{fig:1_shot_factor_correlation_extra}, Appendix). With the high-quality pool, different demonstrations yield similar translation results (see blue points) despite their feature values varying greatly. Several features show insignificant and inconsistent correlation, particularly for De$\rightarrow$En and Zh$\rightarrow$En. This suggests developing selection policy for high-quality example pool is non-trivial.

After mixing with demonstrations from the low-quality pool, the significance gets strengthened. LMScore and CaseSemScore-Tgt shows the highest correlation on average followed by TLength and SemScore. MTScore behaves much worse which might be caused by its instability on sentence-level evaluation~\cite{moghe2022extrinsic}. However, we didn't see significant difference in terms of Spearman's $\rho$ between input-relevant and input-agnostic features~\cite{agrawal2022context}, neither among surface-based, \llm-based or semantic-based features. Surprisingly, the simple feature, S/TLength, yields reasonably high correlation. We argue that long examples could offer \llm with more signals about the task's input and output space. This finding suggests that researchers should select long unlabeled sentences for annotation to improve prompting. Yet, most Spearman's $\rho$s are much smaller than 0.5, indicating a weak/fragile relation.

\begin{figure*}[t]
  \centering
  \small
  \resizebox{\textwidth}{!}
  {\includegraphics[scale=0.50]{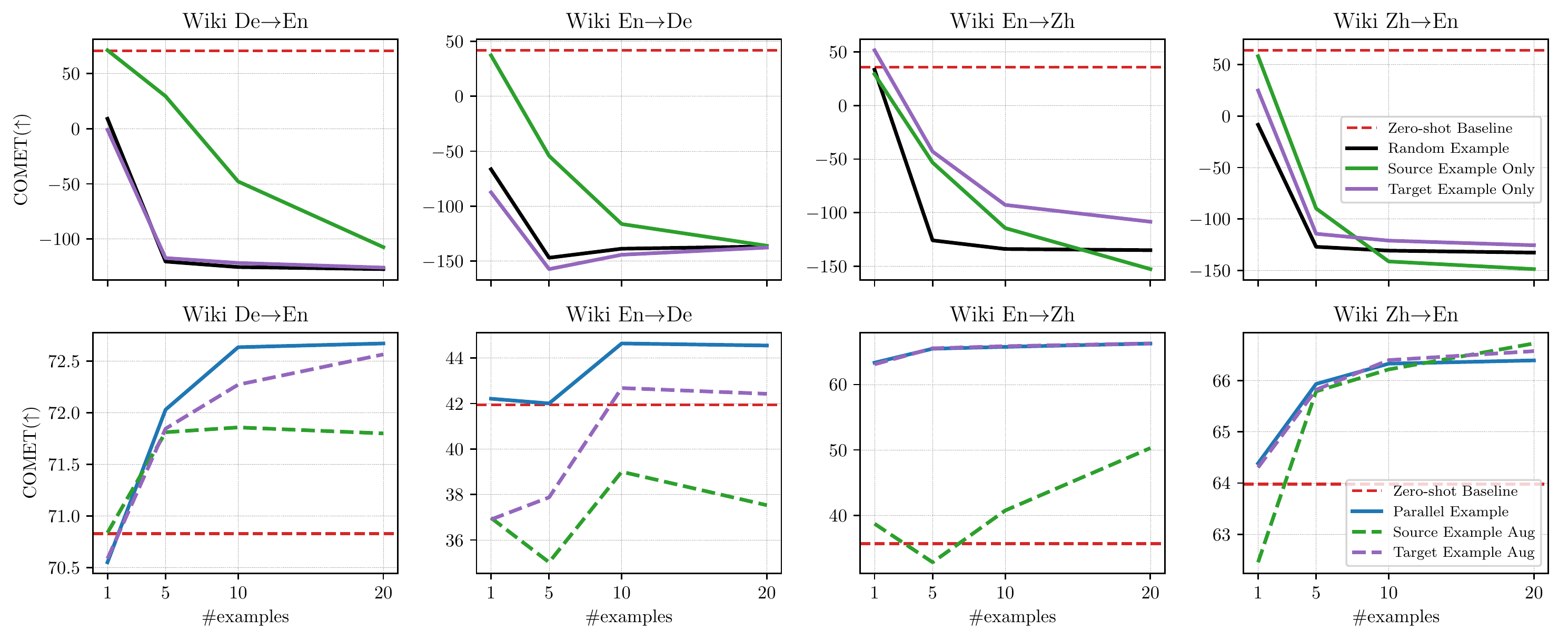}}
  \caption{\label{fig:mono} COMET scores for \textit{few-shot} prompting with monolingual data on Wiki Ablation sets. \textit{Random Example}: random sentence pairs; \textit{Source/Target Example Only}: only use source or target data for prompting; \textit{Source/Target Example Aug}: use pseudo-parallel data instead constructed via zero-shot prompting. For each setup, we randomly sample 50 demonstrations and report average performance.}
\end{figure*}

\textit{In general, selecting prompt examples of high translation quality, high semantic similarity, high \llm likelihood, long sequence length and high similarity to test inputs are all preferable strategies.} Unfortunately, none of them can guarantee optimal translation performance. 

\paragraph{Using prompt examples selected based on the proposed features yields improved performance.} We next verify the above findings on the Full sets. We explore selection strategies based on SemScore, LMScore and TLength (i.e. use top-ranked examples) as they show high average correlation. We didn't analyze CaseSemScore-Tgt as it's more complicated and doesn't make significant difference. Note we excluded too long (more than 100 tokens) or too short (less than 10 tokens) examples during selection. We also consider 5-shot prompting, where we concatenate top-ranked 5 examples in an ascending order~\cite{liu-etal-2022-makes}. 

Table \ref{tab:res_wiki_wmt} shows that, with high-quality pool, adopting the feature-based strategy is likely to outperform the random baseline, and the SemScore-based strategy performs well across different settings (detailed results are available in Table \ref{tab:detailed_res_wiki} and \ref{tab:detailed_res_wmt}, Appendix). These strategies also generalize to 5-shot prompting to some extent. For selection from low-quality pool, we propose a combined strategy: we first choose top-11K examples according to SemScore to filter out poor examples, the top-1K of which are also dropped as they tend to be uninformative (see Table \ref{tab:sem_uninfor} in Appendix); then we re-rank the rest with LMScore and retain top-1K examples, upon which we further apply the TLength-based strategy. In Table \ref{tab:res_wiki_wmt}, this combined strategy outperforms the random one by varying degrees.

\section{Monolingual Data for Prompting}

A longstanding concern in MT is how to utilize unlabeled data to improve translation. While prompting enables few-shot learning reducing the data requirement, exploring whether demonstration could benefit from monolingual examples is still valuable, both for MT study and for understanding of the role of demonstration in prompting. 

\citet{min2022rethinking} argue that the key role of demonstration lies in its support of the input space, the label space and the prompt format, rather than the genuineness of the examples. They found that randomly replacing labels in demonstration barely hurts performance on classification tasks. We reexamine this argument in the context of MT by studying the following three prompting settings: 1) \textit{random examples} constructing sentence pairs from monolingual sources and targets randomly; 2) \textit{source/target example only} using monolingual source/target alone for prompting.

\paragraph{Directly using monolingual data for demonstration doesn't work.} Figure \ref{fig:mono} (top) shows a totally different story (see Figures \ref{fig:mono_extra} and \ref{fig:mono_bleu} in Appendix for more results): monolingual example-based demonstration almost always hurts translation, and the more examples used, the more degeneration yielded. Using random examples misleads the prompting and performs the worst in general; compared to target-only examples, using source examples yields slightly better results except translating into Chinese. This indicates that the genuine source-target mapping should be retained in the demonstration, and also indicates that MT features unique challenges which deserves more attention when studying prompting.

\paragraph{Pseudo parallel examples by forward-/back-translation benefits prompting.} Inspired by data augmentation in MT~\cite{sennrich-etal-2016-improving,zhang2016exploiting}, we next resort to constructing pseudo parallel data. We first adopt \glm to translate the source or target examples via zero-shot prompting, and then use the generated parallel examples as demonstration. Despite low quality, Figure \ref{fig:mono} (bottom) shows that this is an effective way to improve prompting, and using more examples often produces better results. We also observe that back-translation (i.e. translating target monolingual examples) performs better and behaves more robustly than forward-translation (i.e. translating source examples instead), which even approaches prompting with real parallel examples.

\section{Transfer Learning for Prompting}

After obtaining a performant demonstration, we are interested in to what extent its capability could be transferred across different settings, especially from one domain/language pair to another and from sentence-level to document-level translation. While previous studies demonstrate the feasibility with continuous prompts on classification tasks~\cite{wang-etal-2021-transprompt}, transfer for hard prompting on MT has never been investigated. 

Assume that demonstrations $D_1$ and $D_2$ are selected in setting $S_1$ and that $D_1$ performs better (i.e. $D_1 >D_2$), We have the following research questions:
\begin{itemize} 
    \item Could we also expect $D_1 > D_2$ in setting $S_2$?
    \item Whether using demonstrations from $S_1$ could outperform zero-shot prompting in $S_2$?
\end{itemize}
We next study these questions through experiments with 1-shot prompting.

\paragraph{The superiority of a demonstration doesn't generalize across settings.} If the ranking $D_1 > D_2$ holds across settings, the results of the same set of demonstrations in different settings should show high and significant Spearman's correlation. Unfortunately, the correlations in Table \ref{tab:cross_lingual_transfer} and \ref{tab:cross_domain_transfer} are very weak and often insignificant (more results are given in Table \ref{tab:res_cross_lingual_detailed_corr}, \ref{tab:res_cross_lingual_detailed_quality}, and \ref{tab:cross_domain_transfer_bleu}), even for the same language pairs in different directions (\textit{Reversed}) and for similar domains (Wiki$\Rightarrow$WMT). This suggests that we will need setting-specific demonstration to get the optimal translation quality.

\begin{table}[t]
    \centering
    \small
    \setlength{\tabcolsep}{5pt}
    \begin{tabular}{lrrrr}
    \toprule
     \multirow{2}{*}{Setting} & \multicolumn{2}{c}{Correlation} & \multicolumn{2}{c}{$\Delta$ Quality} \\
     \cmidrule(lr){2-3} \cmidrule(lr){4-5} 
     & \multicolumn{1}{c}{BLEU} & \multicolumn{1}{c}{COMET} & \multicolumn{1}{c}{BLEU} & \multicolumn{1}{c}{COMET} \\
    \midrule
    Source Shared & 0.08 & 0.10 & \posres{+0.59} & \posres{+7.03} \\
    Target Shared & 0.20 & 0.24 & \posres{+1.32} & \posres{+9.67} \\
    Reversed & 0.15 & 0.06 & \posres{+1.41} & \posres{+11.56} \\
    \bottomrule
    \end{tabular}
    \caption{\label{tab:cross_lingual_transfer} Spearman's $\rho$ and relative performance for cross-lingual transfer under \textit{1-shot} prompting on Wiki Ablation sets (among En, De and Zh). When studying transfer from language pair $S_1$ to $S_2$, we randomly sample 300 demonstrations from the default pool of $S_1$, and then evaluate them on the Ablation test sets for $S_1$ and $S_2$ respectively, based on which we compute the correlation. The performance is also averaged. \textit{$\Delta$ Quality}: relative quality against the zero-shot baseline. \posres{Blue} cells indicate positive gains. \textit{Source/Target Shared}: average result for transfer settings where the source/target language is shared; \textit{Reversed}: average result for the same language pair but in different directions.}
\end{table}

\begin{table}[t]
    \centering
    \small
    \begin{tabular}{llrrr}
    \toprule
    \multicolumn{2}{l}{Transfer from Wiki to $\Rightarrow$} & \multicolumn{1}{c}{WMT} & \multicolumn{1}{c}{IT} & \multicolumn{1}{c}{Medical} \\
    \midrule
    \multirow{2}{*}{Correlation} & En$\rightarrow$De & \nosig{0.09} & \nosig{0.14} & 0.27$^\ddagger$ \\
    & De$\rightarrow$En & 0.23$^\ddagger$ & 0.20$^\ddagger$ & \nosig{0.13} \\
    \midrule
    \multirow{2}{*}{$\Delta$ Quality} & En$\rightarrow$De & \posres{+4.00} & \posres{+19.52} &  \posres{+7.80} \\
    & De$\rightarrow$En & \posres{+0.10} & \posres{+19.46} & \posres{+1.24} \\
    \bottomrule
    \end{tabular}
    \caption{\label{tab:cross_domain_transfer} Spearman's $\rho$ and relative performance (in COMET) for cross-domain transfer under \textit{1-shot} prompting. We explore transfer from Wiki to Multi-Domain using the Ablation sets. Correlation and performance are calculated in the same way as in cross-lingual transfer, except that we sample 200 demonstrations. $^\ddagger$: statistically significant at $p<0.01$; \nosig{Gray} cells indicate insignificance.}
\end{table}

\begin{table}[t]
    \centering
    \small
    \begin{tabular}{lrrrrr}
    \toprule
    \multirow{1}{*}{Method} & d-BLEU & TC & CP & PT & TCP \\
    \midrule
    Zero-Shot & 30.2 & 47.5 & \textbf{38.7} & 41.6 & 42.4 \\
    \midrule
    SemScore & \textbf{30.5} & \textbf{53.0} & 34.4 & \textbf{43.2} & 42.9 \\
    LMScore & \textbf{30.5} & \textbf{53.0} & 36.8 & 42.9 & \textbf{43.7} \\
    \bottomrule
    \end{tabular}
    \caption{\label{tab:cross_sentence_transfer} Results for transfer learning from sentence-level demonstration to document-level translation under \textit{1-shot} prompting on PDC Zh$\rightarrow$En Full sets. We split each test document in PDC into non-overlapped chunks, each of which contains about 4 sentences. \textit{SemScore/LMScore}: prompt example selection strategy; we apply them to PDC's default pool. We select 3 demonstrations and report average performance. \textit{d-BLEU}: document-level BLEU; \textit{TC/CP/PT/TCP}($\uparrow$): document-specific metrics proposed in ~\cite{sun-etal-2022-rethinking}.}
\end{table}

\paragraph{Using out-of-setting demonstrations can benefit translation.} However, we can still gain from using out-of-setting demonstrations as demonstrated by the positive gains in Table \ref{tab:cross_lingual_transfer} and \ref{tab:cross_domain_transfer}, where we find that transfer in target-shared and reversed settings is relatively easier, and that transfer across distant domains can be successful particularly when in-setting example pool is of low quality. This is also supported by the transfer to document-level translation, where both BLEU and document-specific evaluation get improved as shown in Table \ref{tab:cross_sentence_transfer}. Results in Table \ref{tab:res_multi_domain} show that the transfer is unstable and could deliver negative results, i.e. worse than zero-shot prompting, partially resonating with previous findings~\cite{lin2021few}. We leave the question of how to select prompt examples in transfer learning setups to future.

\section{Discussion}

\begin{table*}[t]
    \centering
    \small
    \begin{tabular}{lp{12cm}}
    \toprule
     \multirow{2}{*}{Source} & \chinese{根据三江源国家公园管理局长江源园区可可西里管理处统计，藏羚羊回迁数量总体呈逐年上升态势，2019年藏羚羊回迁数量为4860只，比2018年增加338只。} \\
     \multirow{3}{*}{Reference} & Statistics from the Sanjiangyuan National Park Administration Yangtze River Origin Park Hoh Xil Management Office show that the number of Tibetan antelopes on the return migration route has been increasing each year, with 4,860 counted in 2019, an increase of 338 over 2018. \\
     \multirow{3}{*}{\glm (1-shot)} & \chinese{According to the\hlred{三江源国家公园管理局长江源园区可可西里管理处}, the total number of re-migration of the Tibetan antelope \uwave{has been on the rise since 2018}, with 4,860 re-migrating in \hlblue{2109}, an increase of 338 compared to \hlblue{2808}. } \\
    \midrule
    \midrule
    \multirow{2}{*}{Prompt in Prompt} & English: Dominic Raab has defended the Government's decision to re-introduce quarantine measures on Spain at short notice. \textbf{Translate from English to Chinese:} Chinese: \\
    \multirow{2}{*}{Reference} & \chinese{针对政府突然做出重新对西班牙实施隔离措施的决定，Dominic Raab 做出了辩解。从英文翻译成中文：} \\
    \multirow{2}{*}{\glm (zero-shot)} & \chinese{多米尼克·拉布(Dominic Raab)对政府决定重新引入西班牙的检疫措施表示支持。\textbf{Translate from English to Chinese:}} \\
    \bottomrule
    \end{tabular}
    \caption{\label{tab:case_study} Case study of translation errors by prompting. Top: copying (in \hlred{red}), mistranslation of date (in \hlblue{blue}), misunderstanding of source (\uwave{wave lines}); Bottom: prompt trap where the model fails to translate the prompt phrase (in \textbf{bold}).}
\end{table*}

\begin{table}[t]
    \centering
    \small
    \begin{tabular}{lrrrr}
    \toprule
    \multirow{2}{*}{Setting} & \multicolumn{2}{c}{0-shot} & \multicolumn{2}{c}{1-shot} \\
    \cmidrule(lr){2-3} \cmidrule(lr){4-5}
    & De$\rightarrow$Zh & Zh$\rightarrow$De & De$\rightarrow$Zh & Zh$\rightarrow$De \\
    \midrule
    Direct & 2.80 & 10.05 & 47.23 & 11.75 \\
    Pivoting & \textbf{19.23} & \textbf{19.53} & \textbf{48.25} & \textbf{25.31} \\
    \bottomrule
    \end{tabular}
    \caption{\label{tab:directxy_vs_pivoting} COMET scores for direct vs. pivoting translation for De$\leftrightarrow$Zh on Wiki Full sets. In 1-shot prompting, we randomly sample 3 demonstrations and report average performance. \textit{Pivoting}: source $\rightarrow$ English $\rightarrow$ target.} 
\end{table}

Although prompting enables translation with decent performance, it still suffers from many (well-known) problems. Here, we briefly explain the problems we observed from the model's outputs.

Prompting sometimes rejects translating the input. Instead, it emits either empty or off-target outputs, i.e. translating in a wrong target language. This occurs frequently when translating into Chinese, where the model often translates into traditional Chinese with messy codes, causing unstable performance. Besides overly relying on a language model, prompting tends to under-translate the input, copy source phrases, produce code-switched output, mistranslate entities (e.g. dates) and generate hallucination, as illustrated in Table \ref{tab:case_study}.

We also observe a phenomenon specific to prompting: \textit{prompt trap} where prompting behaves unpredictable when its input is mixed with prompt template phrases. In the second case in Table \ref{tab:case_study}, the model copies the template phrases, rather than translating them into Chinese. This means that translating prompt itself (not just the input) becomes non-trivial, and that users may attack prompting-based translation systems by manipulating the input format.

We find that the translation quality between German and Chinese is very poor (see Table \ref{tab:detailed_res_wiki}). We argue that the cross-lingual ability of \glm mainly centers around English (although \glm was pretrained on Chinese as well), and thus explore pivoting translation instead. Table \ref{tab:directxy_vs_pivoting} shows that pivoting through English greatly improves non-English translation. It's still unclear whether the current \llm pretraining recipe could achieve promising non-English-centric cross-lingual ability. We might need to consider adding parallel data into the \llm pretraining or finetuning.

\section{Related Work}

The capability of prompting heavily depends on its surface representation, where small modifications to the prompt could cause high variance in its performance. This inspires researchers to develop advanced prompting strategies to get the most from \llms. \citet{gao-etal-2021-making} proposed to generate prompt templates automatically using T5~\cite{xue2021mt5} rather than adopting manual templates. \citet{liu-etal-2022-makes} reported selecting prompt examples close to the test input via a $k$NN-based retriever, \citet{sorensen-etal-2022-information} resorted to an information-theoretic approach based on mutual information, while \citet{zhang2022active} formulated example selection as a sequential decision problem and solved it by reinforcement learning. For reasoning tasks, \citet{wei2022chain} developed chain-of-thought (CoT) prompting letting the model output the intermediate reasoning steps, which inspires researchers to further explore CoT selection~\cite{fu2022complexity} and decomposition~\cite{zhou2022least}. In contrast to the studies just mentioned, which focus on NLP tasks other than MT, we explore prompting strategies exclusively for translation.

Prompting uses instructions to guide \llms, which is closely related to neural MT with special prefixes. In multilingual NMT, a target language tag is often appended to the source input to indicate the translation direction~\cite{TACL1081,arivazhagan2019massively,zhang-etal-2020-improving}. Special attribute tags can also be used to control properties of the model output, such as politeness~\cite{sennrich-etal-2016-controlling}, diversity~\cite{shu-etal-2019-generating}, and quality~\cite{caswell-etal-2019-tagged}. Besides, retrieved phrases and sentences can be augmented to the input to improve translation quality~\cite{zhang-etal-2018-guiding,10.5555/3504035.3504664}. With the popularity of prompting \llms, researchers see value in incorporating prompts into neural MT~\cite{li-etal-2022-prompt,tan2021msp,garcia2022using}. Still, these methods rely on pretraining or finetuning the model rather than prompting frozen \llms.

Very recently, concurrent to our work, \citet{vilar2022prompting} examined the capability of prompting PaLM for translation and discovered that prompting with high-quality examples even chosen randomly performs on par with or better than the one using input-relevant examples. By contrast, \citet{agrawal2022context} explored strategies to select input-specific examples, and observed that input-relevant examples based on n-gram overlap significantly improves the capability of prompts. Our study resonates with both their findings and also explains their conflict: while the quality and input-based semantic similarity correlate with prompting performance significantly, the correlation strength is unfortunately not strong enough so using them as indicators to select examples may produce mixed results. Note that apart from example selection, we also studied using monolingual data and transfer learning for MT prompting, which, to the best of our knowledge, have never been explored before.

\section{Conclusion and Future Work}

In this paper, we  presented a systematic study on prompting for MT, exploring topics ranging from prompting strategy, the use of unlabelled monolingual data, to transfer learning.  We found that prompt template and demonstration example selection both have substantial impact on translation. Some prompt example features correlate significantly with prompting performance; treating them as criteria for example selection benefits translation to some extent but not consistently as the correlations are not strong enough.

Prompting for MT requires retaining the source-target mapping signals in the demonstration. Directly applying monolingual data for prompting sounds interesting but doesn't work. Constructing pseudo parallel prompt examples by back-/forward-translation via zero-shot prompting is a simple yet effective solution. Regarding transfer learning, we saw positive results when applying a (sentence-level) demonstration to other domains, other language pairs or document-level translation. Unfortunately, the optimality of the demonstration doesn't generalize across settings and the transfer performance is also unstable. We argue that MT provides a set of unique challenges and call for more efforts on evaluating prompting \llms for MT.

Prompting also faces a number of other issues, like off-target generation and prompt traps, which we plan to address in the future. We are also interested in examining whether our findings can generalize to other \llms, like GPT-3, OPT and PaLM. We would also like to explore further how to improve the cross-lingual ability in \llm.


\section*{Limitations}

Our study heavily depends on the INT-4 quantized \glm, which, unlike GPT and PaLM, was pretrained with both bidirectional and unidirectional training objectives. The quantization might weaken the model's capability and deteriorate some unknown aspects. It's unclear how our findings generalize to other pretrained \llms. In addition, we mainly work on three languages due to resource constraints, and in experiments, results vary greatly across language pairs. Increasing the coverage of experimental languages would make the results more reliable.

\section*{Acknowledgments}
This work was funded by UK Research and Innovation (UKRI) under the UK government’s Horizon Europe funding guarantee [grant number 10039436 -- UTTER]. The computations described in this research were performed using the Baskerville Tier 2 HPC service (https://www.baskerville.ac.uk/). Baskerville was funded by the EPSRC and UKRI through the World Class Labs scheme (EP/T022221/1) and the Digital Research Infras-tructure programme (EP/W032244/1) and is operated by Advanced Research Computing at the University of Birmingham.


\bibliography{paper}
\bibliographystyle{acl_natbib}


\appendix

\section{Appendix} \label{sec:appendix}

\begin{table*}[t]
    \centering
    \setlength{\tabcolsep}{4pt}
    
    \subcaptionbox{\label{tab:data_statistics_ablation} Ablation Sets}{
       \small
        \begin{tabular}{llccc}
        \toprule
        Dataset & Language(s) & Test Set & Selection Pool (Default) & Source (\#sample) \\
        \midrule
        \multirow{3}{*}{Wiki} & English & 100 & 897 & \flores \texttt{eng\_Latn.dev} (997) \\
                              & German & 100  & 897 & \flores \texttt{deu\_Latn.dev} (997) \\
                              & Chinese & 100 & 897 & \flores \texttt{zho\_Hans.dev} (997) \\
        \midrule
        WMT & English-German & 100 & 2900 & newstest2013 (3000) \\
        \midrule
        IT & German-English & 100 & 1900 & Multi-Domain Dev Set (2000) \\
        Medical & German-English & 100 & 1900 & Multi-Domain Dev Set (2000) \\
        \bottomrule
        \end{tabular}
    }
    
    \vspace{\baselineskip}
    \subcaptionbox{\label{tab:data_statistics_full} Full Sets}{
        \small
        \begin{tabular}{llcccc}
        \toprule
        Dataset & Languages & Source &  Test Set & High-quality Pool (Default) & Low-quality Pool \\
        \midrule
        \multirow{3}{*}{Wiki} & English & \flores & \texttt{eng\_Latn.devtest} (1012) & \texttt{eng\_Latn.dev} (997) & \multirow{3}{*}{\makecell{
                                                En-Zh$^\star$ (0.79M) \\ De-En$^\star$ (1.57M) \\ De-Zh$^\star$ (0.13M)}} \\
                              & German & \flores &  \texttt{deu\_Latn.devtest} (1012) &  \texttt{deu\_Latn.dev} (997)   \\
                              & Chinese & \flores &  \texttt{zho\_Hans.devtest} (1012) &  \texttt{zho\_Hans.dev} (997) \\
        \midrule
        \multirow{2}{*}{WMT} & English-German & WMT & newstest2021 (1002/1000) & newstest2020 (1418) \\
                            & English-Chinese & WMT & newstest2021 (1002/1948) & newstest2020 (1418) \\
        \midrule
        IT & German-English & Multi-Domain & Test Set (2000) & - & Train Set (0.22M) \\
        Law & German-English & Multi-Domain & Test Set (2000) & - & Train Set (0.47M) \\
        Medical & German-English & Multi-Domain &  Test Set (2000) & - & Train Set (0.25M) \\
        \midrule
        PDC & Chinese-English & News & Test Set (4858/148 Docs) & Dev Set (2881) & - \\
        \bottomrule
        \end{tabular}
    }
    
    \caption{\label{tab:data_statistics} Statistics of Ablation sets and Full sets. Numbers in brackets denote the number of instances. $^\star$: data from WikiMatrix.v1~\cite{schwenk-etal-2021-wikimatrix}.}
\end{table*}

\begin{table*}[t]
    \centering
    \small
    \setlength{\tabcolsep}{3.8pt}
    \begin{tabular}{lrrrrrrrrrrrrrr}
    \toprule
    \multirow{4}{*}{ID} & \multicolumn{7}{c}{BLEU} & \multicolumn{7}{c}{COMET} \\
    \cmidrule(lr){2-8} \cmidrule(lr){9-15} 
    & \multicolumn{2}{c}{De $\leftrightarrow$ En} & \multicolumn{2}{c}{De $\leftrightarrow$ Zh} & \multicolumn{2}{c}{En $\leftrightarrow$ Zh} & \multirow{2}{*}{Avg} & \multicolumn{2}{c}{De $\leftrightarrow$ En} & \multicolumn{2}{c}{De $\leftrightarrow$ Zh} & \multicolumn{2}{c}{En $\leftrightarrow$ Zh} & \multirow{2}{*}{Avg} \\
    \cmidrule(lr){2-3} \cmidrule(lr){4-5} \cmidrule(lr){6-7}
    \cmidrule(lr){9-10} \cmidrule(lr){11-12} \cmidrule(lr){13-14}
    & \multicolumn{1}{c}{$\rightarrow$} & \multicolumn{1}{c}{$\leftarrow$} & \multicolumn{1}{c}{$\rightarrow$} & \multicolumn{1}{c}{$\leftarrow$} & \multicolumn{1}{c}{$\rightarrow$} & \multicolumn{1}{c}{$\leftarrow$} & & \multicolumn{1}{c}{$\rightarrow$} & \multicolumn{1}{c}{$\leftarrow$} & \multicolumn{1}{c}{$\rightarrow$} & \multicolumn{1}{c}{$\leftarrow$} & \multicolumn{1}{c}{$\rightarrow$} & \multicolumn{1}{c}{$\leftarrow$} \\
    \midrule
    \multicolumn{10}{l}{\it English Template \textbf{Without} Line Break}  \vspace{0.1cm} \\
    A & \textbf{38.00} & \textbf{23.10} & 23.30 & \textbf{12.10} & \underline{31.50} & \textbf{27.90} & \textbf{25.98} & \textbf{70.83} & \textbf{41.95} & 4.34 & \textbf{15.92} & \underline{35.68} & \textbf{63.98} & \textbf{38.78} \\
    B & 8.30 & 9.00 & 2.80 & 2.40 & 6.60 & 8.20 & 6.22 & -45.75 & -70.27 & -140.43 & -119.82 & -112.38 & -43.10 & -88.62 \\
    C & 30.60 & 2.10 & 5.50 & 1.10 & 1.10 & 8.30 & 8.12 & 29.78 & -142.36 & -117.20 & -117.14 & -120.57 & -58.32 & -87.63 \\
    D & 26.10 & 0.00 & 5.10 & 0.00 & 0.20 & 0.60 & 5.33 & -1.20 & -160.59 & -124.15 & -157.62 & -130.51 & -108.71 & -113.80 \\
    E & 35.90 & 18.20 & \underline{26.10} & 9.60 & 16.00 & 22.30 & 21.35 & 68.06 & 5.41 & \underline{27.53} & -6.46 & -5.58 & 35.93 & 20.81 \\
    F & 33.50 & 5.60 & 25.10 & 0.80 & 0.20 & 9.10 & 12.38 & 61.09 & -62.31 & 22.71 & -112.79 & -50.84 & -20.71 & -27.14 \\
    \midrule
    \multicolumn{10}{l}{\it English Template \textbf{With} Line Break}  \vspace{0.1cm} \\
    A & \underline{36.60} & \underline{21.80} & \underline{25.10} & \underline{11.40} & 26.90 & \underline{26.90} & \underline{24.78} & \underline{67.97} & \underline{37.41} & 7.24 & \underline{9.46} & 4.89 & \underline{60.08} & \underline{31.17} \\
    B & 7.70 & 7.70 & 5.00 & 2.70 & 13.20 & 10.00 & 7.72 & -85.97 & -81.79 & -126.58 & -113.27 & -55.64 & -48.82 & -85.35 \\
    C & 28.00 & 4.40 & 7.70 & 0.70 & 13.30 & 14.50 & 11.43 & 36.10 & -99.01 & -118.99 & -133.39 & -74.19 & -23.00 & -68.75 \\
    D & 25.20 & 1.60 & 4.20 & 0.10 & 4.90 & 5.40 & 6.90 & 13.96 & -121.58 & -125.36 & -148.29 & -78.78 & -74.91 & -89.16 \\
    E & 35.70 & 20.00 & 24.40 & 3.90 & \underline{28.30} & 20.30 & 22.10 & 66.08 & 22.21 & \underline{15.62} & -55.41 & \underline{13.36} & 38.30 & 16.69 \\
    F & 33.60 & 9.30 & 23.60 & 3.00 & 6.70 & 17.90 & 15.68 & 57.46 & -45.84 & 14.73 & -69.69 & -30.63 & 32.68 & -6.88 \\
    \midrule
    \midrule
    \multicolumn{10}{l}{\it German Template \textbf{Without} Line Break}  \vspace{0.1cm} \\
    A & \underline{20.00} & 15.70 & \underline{1.60} & 3.10 & 0.70 & 7.10 & 8.03 & \underline{23.09} & 4.61 & -70.84 & -47.51 & -65.61 & -0.66 & -26.15 \\
    B & 5.60 & 2.10 & 0.10 & 1.60 & 0.20 & 1.10 & 1.78 & -82.99 & -152.26 & -174.72 & -132.06 & -162.79 & -110.99 & -135.97 \\
    C & 4.60 & 5.40 & 0.30 & 3.70 & 0.00 & 4.10 & 3.02 & -57.63 & -108.36 & -120.99 & -125.18 & -135.21 & -90.42 & -106.30 \\
    D & 3.50 & 0.10 & 0.00 & 0.00 & 0.00 & 0.10 & 0.62 & -115.55 & -168.13 & -166.07 & -169.21 & -161.27 & -142.57 & -153.80 \\
    E & 17.30 & \underline{19.00} & 0.20 & \underline{8.50} & \underline{2.30} & \underline{19.60} & \underline{11.15} & 14.19 & \underline{6.47} & -100.92 & \underline{-25.14} & \underline{-50.42} & \underline{9.85} & \underline{-24.33} \\
    F & 6.30 & 4.80 & 0.20 & 7.30 & 0.10 & 11.70 & 5.07 & 3.88 & -65.86 & \underline{-44.76} & -27.91 & -60.31 & -11.22 & -34.36 \\
    \midrule
    \multicolumn{10}{l}{\it German Template \textbf{With} Line Break}  \vspace{0.1cm} \\
    A & \underline{25.40} & \underline{20.20} & 6.40 & 3.50 & 8.00 & 9.20 & 12.12 & \underline{38.47} & \underline{31.45} & -80.14 & -47.22 & -50.26 & 8.84 & -16.48 \\
    B & 15.60 & 7.80 & 2.60 & 1.00 & 0.50 & 0.80 & 4.72 & -20.65 & -81.28 & -125.21 & -137.02 & -125.31 & -108.45 & -99.65 \\
    C & 15.40 & 5.70 & 5.70 & 3.00 & 6.00 & 6.70 & 7.08 & -23.46 & -80.15 & -86.27 & -104.10 & -87.18 & -58.23 & -73.23 \\
    D & 2.80 & 0.50 & 0.00 & 0.00 & 0.10 & 1.10 & 0.75 & -95.30 & -154.76 & -140.51 & -155.91 & -137.36 & -100.08 & -130.65 \\
    E & 24.70 & 19.50 & \underline{10.40} & 8.50 & \underline{11.10} & \underline{17.20} & \underline{15.23} & 35.12 & 3.95 & -62.48 & -18.32 & \underline{-27.61} & \underline{35.26} & \underline{-5.68} \\
    F & 7.60 & 17.20 & 0.50 & \underline{8.60} & 3.90 & 11.30 & 8.18 & 13.01 & 9.10 & \underline{-43.63} & \underline{-10.88} & -46.46 & 23.54 & -9.22 \\
    \midrule
    \midrule
    \multicolumn{10}{l}{\it Chinese Template \textbf{Without} Line Break}  \vspace{0.1cm} \\
    A & \underline{37.60} & \underline{15.50} & \textbf{28.30} & \underline{2.10} & 33.40 & 15.10 & \underline{22.00} & \underline{67.41} & \underline{-5.40} & \textbf{45.24} & \underline{-74.78} & 53.71 & 2.72 & \underline{14.82} \\
    B & 23.60 & 6.30 & 14.50 & 0.50 & 19.30 & 1.90 & 11.02 & -6.41 & -90.63 & -12.10 & -159.66 & -9.24 & -121.29 & -66.55 \\
    C & 11.40 & 3.20 & 14.30 & 0.40 & 20.80 & 5.00 & 9.18 & -32.55 & -114.57 & -9.91 & -140.54 & 2.89 & -85.58 & -63.38 \\
    D & 17.10 & 6.40 & 15.90 & 0.20 & 19.60 & 1.90 & 10.18 & -34.15 & -101.69 & -24.36 & -166.15 & -9.20 & -125.20 & -76.79 \\
    E & 29.00 & 8.00 & 27.00 & 0.40 & \textbf{34.90} & \underline{16.10} & 19.23 & 35.55 & -63.09 & 37.06 & -119.13 & \textbf{54.14} & \underline{3.80} & -8.61 \\
    F & 31.70 & 3.70 & 24.80 & 0.10 & 27.20 & 11.80 & 16.55 & 35.65 & -105.74 & 22.97 & -129.71 & 5.61 & -34.09 & -34.22  \\
    \midrule
    \multicolumn{10}{l}{\it Chinese Template \textbf{With} Line Break}  \vspace{0.1cm} \\
    A & 26.80 & \underline{14.70} & 24.70 & \underline{3.30} & \underline{33.80} & \underline{22.90} & \underline{21.03} & \underline{24.46} & \underline{-84.74} & \underline{24.76} & \underline{-64.07} & \underline{52.65} & \underline{40.45} & \underline{-1.08} \\
    B & 23.70 & 6.30 & 11.90 & 0.10 & 14.40 & 0.60 & 9.50 & -11.65 & -102.50 & -63.95 & -161.96 & -46.84 & -128.12 & -85.84 \\
    C & 12.10 & 3.00 & 13.80 & 0.80 & 21.20 & 9.90 & 10.13 & -36.39 & -105.55 & -42.16 & -151.06 & -15.41 & -74.90 & -70.91 \\
    D & 14.10 & 3.20 & 15.10 & 0.20 & 20.00 & 2.50 & 9.18 & -19.15 & -106.69 & -19.34 & -154.73 & -11.51 & -94.82 & -67.71 \\
    E & \underline{28.60} & 8.00 & \underline{26.50} & 0.90 & 32.30 & 21.40 & 19.62 & 8.71 & -118.14 & 15.34 & -124.30 & 21.18 & 14.91 & -30.38 \\
    F & 26.90 & 3.40 & 26.10 & 0.20 & 25.80 & 16.00 & 16.40 & 11.58 & -120.31 & 10.33 & -129.61 & -21.19 & -20.52 & -44.95 \\
    \bottomrule
    \end{tabular}
    \caption{\label{tab:res_template_detailed_comet} Detailed \textit{zero-shot} results for prompting with different templates and different template languages on Wiki Ablation sets. Template \textcircled{A} in English achieves the overall best performance measured by BLEU and COMET. \textit{Avg}: average result over different language pairs. Best results in each section are \underline{underlined}; best results in each column are in \textbf{bold}.}
\end{table*}

\begin{figure*}[t]
  \centering
  \small
  \resizebox{\textwidth}{!}{\includegraphics[scale=0.50]{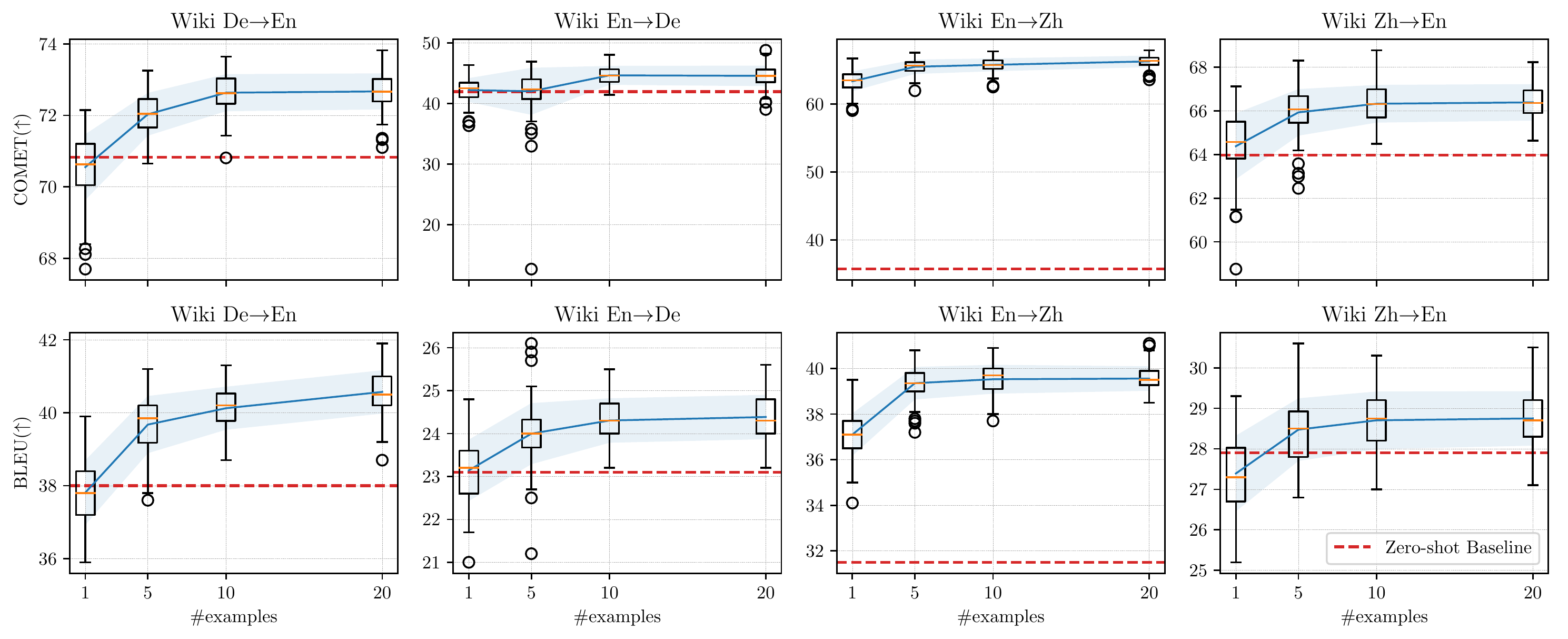}}
  \resizebox{\columnwidth}{!}{\includegraphics[scale=0.50]{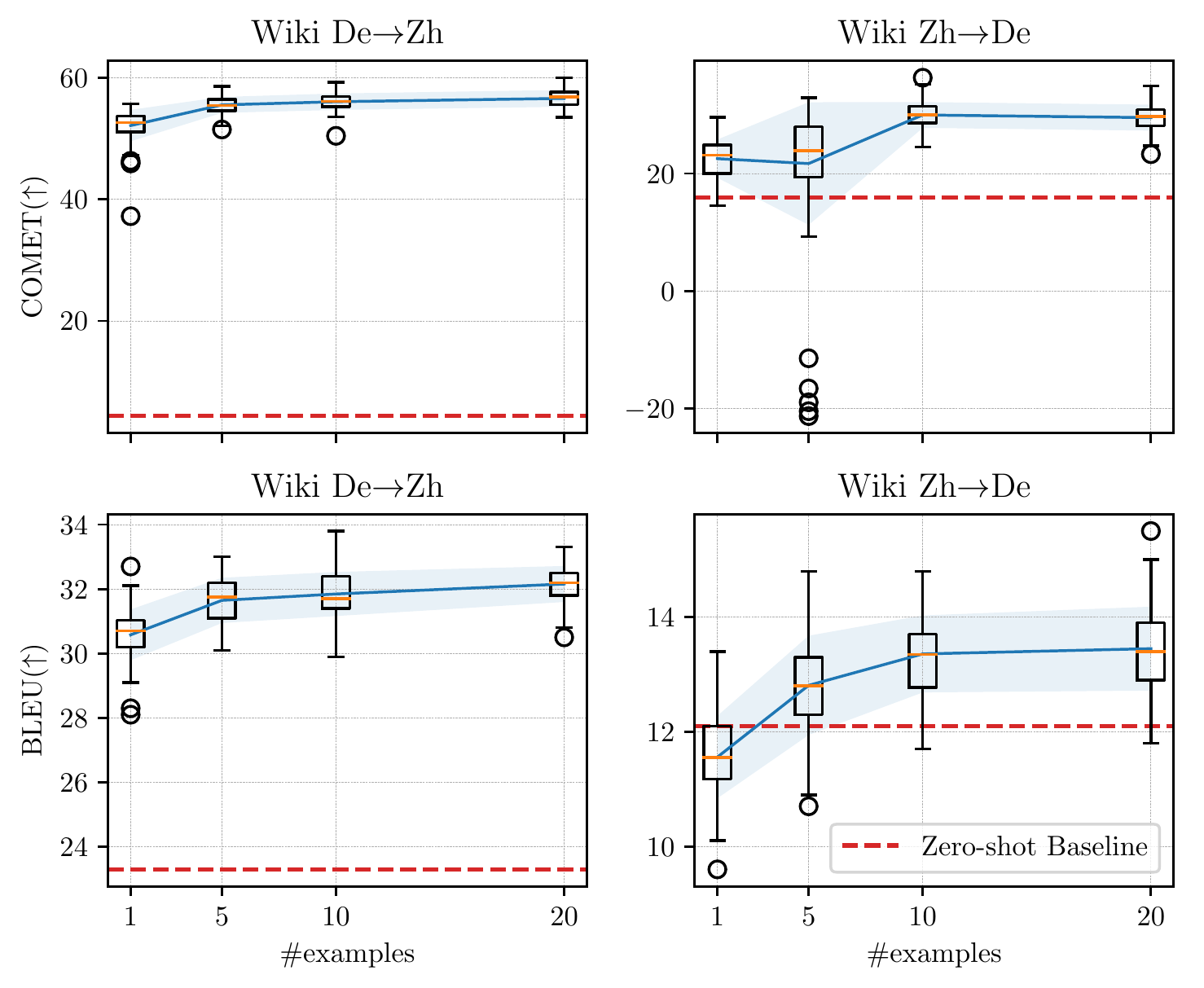}}
  \caption{\label{fig:num_examples_extra} COMET (top) and BLEU (bottom) scores for \textit{few-shot} prompting as a function of the number of prompt examples ($K=1,5,10,20$) on Wiki Ablation sets. For each setup, we randomly sample 100 times from the example pool and show the performance distribution via box plots. Dashed red line denotes the zero-shot baseline; blue curve and shadow area denote the mean and standard deviation.}
\end{figure*}


\begin{table*}[t]
    \centering
    \small
    \setlength{\tabcolsep}{2.8pt}
    \begin{tabular}{lrrrrrrrrrrrrrr}
    \toprule
    \multirow{3}{*}{Method} & \multicolumn{7}{c}{High-quality Examples} & \multicolumn{7}{c}{Plusll Low-quality Examples} \\
    \cmidrule(lr){2-8} \cmidrule(lr){9-15} 
    & \multicolumn{2}{c}{De $\leftrightarrow$ En} & \multicolumn{2}{c}{De $\leftrightarrow$ Zh} & \multicolumn{2}{c}{En $\leftrightarrow$ Zh} & \multirow{2}{*}{Avg} & \multicolumn{2}{c}{De $\leftrightarrow$ En} & \multicolumn{2}{c}{De $\leftrightarrow$ Zh} & \multicolumn{2}{c}{En $\leftrightarrow$ Zh} & \multirow{2}{*}{Avg} \\
    \cmidrule(lr){2-3} \cmidrule(lr){4-5} \cmidrule(lr){6-7}
    \cmidrule(lr){9-10} \cmidrule(lr){11-12} \cmidrule(lr){13-14}
    & \multicolumn{1}{c}{$\rightarrow$} & \multicolumn{1}{c}{$\leftarrow$} & \multicolumn{1}{c}{$\rightarrow$} & \multicolumn{1}{c}{$\leftarrow$} & \multicolumn{1}{c}{$\rightarrow$} & \multicolumn{1}{c}{$\leftarrow$} & & \multicolumn{1}{c}{$\rightarrow$} & \multicolumn{1}{c}{$\leftarrow$} & \multicolumn{1}{c}{$\rightarrow$} & \multicolumn{1}{c}{$\leftarrow$} & \multicolumn{1}{c}{$\rightarrow$} & \multicolumn{1}{c}{$\leftarrow$} \\
    \midrule
    \multicolumn{10}{l}{\textit{Correlation with COMET}} \vspace{0.1cm} \\
    SLength & \nosig{ 0.02} & \lessthree{ 0.18}$^\ddagger$ & \lessthree{ 0.24}$^\ddagger$ & \lessthree{ 0.12}$^\ddagger$ & \lessthree{ 0.26}$^\ddagger$ & \nosig{ 0.01} &  0.14  & \lessthree{ 0.09}$^\ddagger$ & \lessthree{ 0.20}$^\ddagger$ & \geaterfive{ 0.52}$^\ddagger$ & { 0.44}$^\ddagger$ & \lessthree{ 0.24}$^\ddagger$ & \lessthree{ 0.10}$^\ddagger$ &  0.26 \\
    TLength & \nosig{-0.01} & \lessthree{ 0.23}$^\ddagger$ & \lessthree{ 0.19}$^\ddagger$ & \lessthree{ 0.27}$^\ddagger$ & \lessthree{ 0.29}$^\ddagger$ & \nosig{ 0.06} &  \textbf{0.17}  & \lessthree{ 0.06}$^\dagger$ & { 0.35}$^\ddagger$ & { 0.41}$^\ddagger$ & \geaterfive{ 0.57}$^\ddagger$ & \lessthree{ 0.25}$^\ddagger$ & \lessthree{ 0.13}$^\ddagger$ &  0.29 \\
    LMScore & \nosig{ 0.06} & \lessthree{ 0.23}$^\ddagger$ & \nosig{ 0.01} & \lessthree{ 0.20}$^\ddagger$ & \lessthree{ 0.12}$^\ddagger$ & \lessthree{ 0.21}$^\ddagger$ &  0.14  & \lessthree{ 0.19}$^\ddagger$ & { 0.38}$^\ddagger$ & { 0.35}$^\ddagger$ & \geaterfive{ 0.51}$^\ddagger$ & \lessthree{ 0.16}$^\ddagger$ & \lessthree{ 0.27}$^\ddagger$ &  \textbf{0.31} \\
    MTScore & \nosig{ 0.01} & \nosig{ 0.05} & \lessthree{ 0.11}$^\ddagger$ & \lessthree{ 0.12}$^\ddagger$ & \nosig{ 0.06} & \lessthree{ 0.28}$^\ddagger$ &  0.11 & \lessthree{ 0.13}$^\ddagger$ & \nosig{ 0.04} & { 0.30}$^\ddagger$ & \lessthree{ 0.23}$^\ddagger$ & \lessthree{ 0.18}$^\ddagger$ & \lessthree{ 0.28}$^\ddagger$ &  0.19 \\
    SemScore & \lessthree{ 0.11}$^\ddagger$ & \lessthree{ 0.17}$^\ddagger$ & \lessthree{ 0.11}$^\ddagger$ & \lessthree{ 0.15}$^\ddagger$ & \lessthree{ 0.10}$^\ddagger$ & { 0.31}$^\ddagger$ &  0.16  & \lessthree{ 0.12}$^\ddagger$ & \lessthree{ 0.24}$^\ddagger$ & { 0.42}$^\ddagger$ & { 0.50}$^\ddagger$ & \lessthree{ 0.17}$^\ddagger$ & { 0.33}$^\ddagger$ &  0.30 \\
    CaseSemScore-Src & \nosig{-0.01} & \lessthree{ 0.20}$^\ddagger$ & \lessthree{ 0.22}$^\ddagger$ & \lessthree{ 0.08}$^\dagger$ & \lessthree{ 0.18}$^\ddagger$ & \nosig{-0.03} &  0.11 & \lessthree{ 0.08}$^\ddagger$ & \lessthree{ 0.29}$^\ddagger$ & \geaterfive{ 0.53}$^\ddagger$ & { 0.49}$^\ddagger$ & \lessthree{ 0.26}$^\ddagger$ & \nosig{ 0.05} &  0.28 \\
    CaseSemScore-Tgt & \nosig{-0.01} & \lessthree{ 0.22}$^\ddagger$ & \lessthree{ 0.25}$^\ddagger$ & \lessthree{ 0.14}$^\ddagger$ & \lessthree{ 0.21}$^\ddagger$ & \nosig{ 0.05} &  0.14  & \lessthree{ 0.09}$^\ddagger$ & { 0.32}$^\ddagger$ & \geaterfive{ 0.53}$^\ddagger$ & \geaterfive{ 0.53}$^\ddagger$ & \lessthree{ 0.27}$^\ddagger$ & \lessthree{ 0.11}$^\ddagger$ &  \textbf{0.31} \\
    \midrule
    \multicolumn{10}{l}{\textit{Correlation with BLEU}} \vspace{0.1cm} \\
    SLength & \lessthree{ 0.20}$^\ddagger$ & \lessthree{ 0.27}$^\ddagger$ & \lessthree{ 0.21}$^\ddagger$ & \lessthree{ 0.11}$^\ddagger$ & { 0.33}$^\ddagger$ & \lessthree{ 0.12}$^\ddagger$ &  0.21  & \lessthree{ 0.23}$^\ddagger$ & \lessthree{ 0.30}$^\ddagger$ & \geaterfive{ 0.51}$^\ddagger$ & { 0.35}$^\ddagger$ & \lessthree{ 0.29}$^\ddagger$ & \lessthree{ 0.18}$^\ddagger$ &  0.31 \\
    TLength & \lessthree{ 0.15}$^\ddagger$ & { 0.32}$^\ddagger$ & \lessthree{ 0.16}$^\ddagger$ & \lessthree{ 0.22}$^\ddagger$ & { 0.40}$^\ddagger$ & \lessthree{ 0.12}$^\ddagger$ &  \textbf{0.23} & \lessthree{ 0.15}$^\ddagger$ & { 0.38}$^\ddagger$ & { 0.41}$^\ddagger$ & { 0.47}$^\ddagger$ & { 0.33}$^\ddagger$ & \lessthree{ 0.19}$^\ddagger$ &  {0.32} \\
    LMScore & \lessthree{ 0.14}$^\ddagger$ & \lessthree{ 0.17}$^\ddagger$ & \lessthree{ 0.10}$^\ddagger$ & \lessthree{ 0.24}$^\ddagger$ & \lessthree{ 0.27}$^\ddagger$ & \lessthree{ 0.26}$^\ddagger$ &  0.20 & \lessthree{ 0.23}$^\ddagger$ & \lessthree{ 0.30}$^\ddagger$ & { 0.39}$^\ddagger$ & { 0.46}$^\ddagger$ & \lessthree{ 0.27}$^\ddagger$ & { 0.32}$^\ddagger$ &  \textbf{0.33} \\
    MTScore & \nosig{ 0.03} & \nosig{-0.05} & \nosig{ 0.04} & \lessthree{ 0.09}$^\dagger$ & \nosig{ 0.03} & \lessthree{ 0.12}$^\ddagger$ &  0.04 & \lessthree{ 0.11}$^\ddagger$ & \nosig{-0.04} & \lessthree{ 0.26}$^\ddagger$ & \lessthree{ 0.19}$^\ddagger$ & \lessthree{ 0.17}$^\ddagger$ & \lessthree{ 0.14}$^\ddagger$ &  0.14 \\
    SemScore & \lessthree{ 0.13}$^\ddagger$ & \lessthree{ 0.11}$^\ddagger$ & \lessthree{ 0.15}$^\ddagger$ & \lessthree{ 0.20}$^\ddagger$ & \lessthree{ 0.25}$^\ddagger$ & \lessthree{ 0.29}$^\ddagger$ &  0.19  & \lessthree{ 0.13}$^\ddagger$ & \lessthree{ 0.20}$^\ddagger$ & { 0.45}$^\ddagger$ & { 0.45}$^\ddagger$ & \lessthree{ 0.28}$^\ddagger$ & { 0.31}$^\ddagger$ &  0.30 \\
    CaseSemScore-Src & \lessthree{ 0.16}$^\ddagger$ & \lessthree{ 0.15}$^\ddagger$ & \lessthree{ 0.18}$^\ddagger$ & \nosig{ 0.03} & \lessthree{ 0.28}$^\ddagger$ & \nosig{ 0.03} &  0.14  & \lessthree{ 0.20}$^\ddagger$ & \lessthree{ 0.29}$^\ddagger$ & \geaterfive{ 0.51}$^\ddagger$ & { 0.36}$^\ddagger$ & { 0.31}$^\ddagger$ & \lessthree{ 0.07}$^\ddagger$ &  0.29 \\
    CaseSemScore-Tgt & \lessthree{ 0.14}$^\ddagger$ & \lessthree{ 0.17}$^\ddagger$ & \lessthree{ 0.16}$^\ddagger$ & \nosig{ 0.05} & \lessthree{ 0.24}$^\ddagger$ & \lessthree{ 0.09}$^\dagger$ &  0.14 & \lessthree{ 0.18}$^\ddagger$ & { 0.30}$^\ddagger$ & { 0.49}$^\ddagger$ & { 0.39}$^\ddagger$ & \lessthree{ 0.29}$^\ddagger$ & \lessthree{ 0.13}$^\ddagger$ &  0.30 \\
    \bottomrule
    \end{tabular}
    \caption{\label{tab:1_shot_factor_correlation_bleu} Detailed Spearman's $\rho$ between demonstration features and their prompting performance (COMET and BLEU) for \textit{1-shot} prompting on Wiki Ablation sets. We randomly sample 600 demonstrations from each pool to calculate the correlation. \textit{High-quality examples} are from the default selection pool while \textit{Low-quality examples} are from WikiMatrix.v1. $^\dagger/^\ddagger$: statistically significant at $p<0.05/0.01$. \nosig{Gray} cells indicate insignificance; \geaterfive{Red} cells indicate $\rho > 0.5$.}
\end{table*}

\begin{figure*}[t]
  \centering
  \small
  \resizebox{\textwidth}{!}{\includegraphics[scale=0.50]{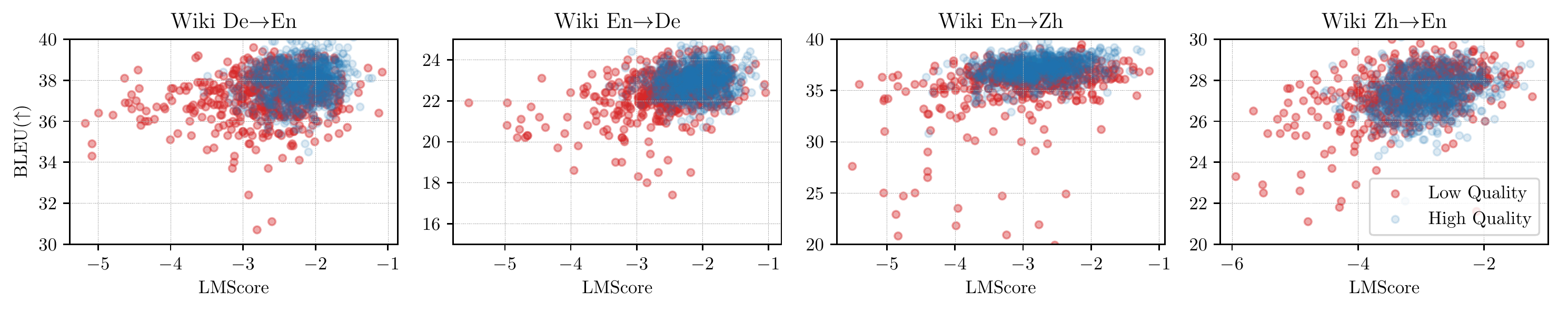}}
  \caption{\label{fig:1_shot_factor_correlation_bleu} Scatter plotting between BLEU and LMScore for \textit{1-shot} prompting on Wiki De$\leftrightarrow$En, En$\leftrightarrow$Zh Ablation sets.}
\end{figure*}

\begin{figure*}[t]
    \centering
    \subcaptionbox{\label{fig:1_shot_factor_correlation_comet_extra} COMET vs. LMScore}{
        \resizebox{\columnwidth}{!}{\includegraphics[scale=0.50]{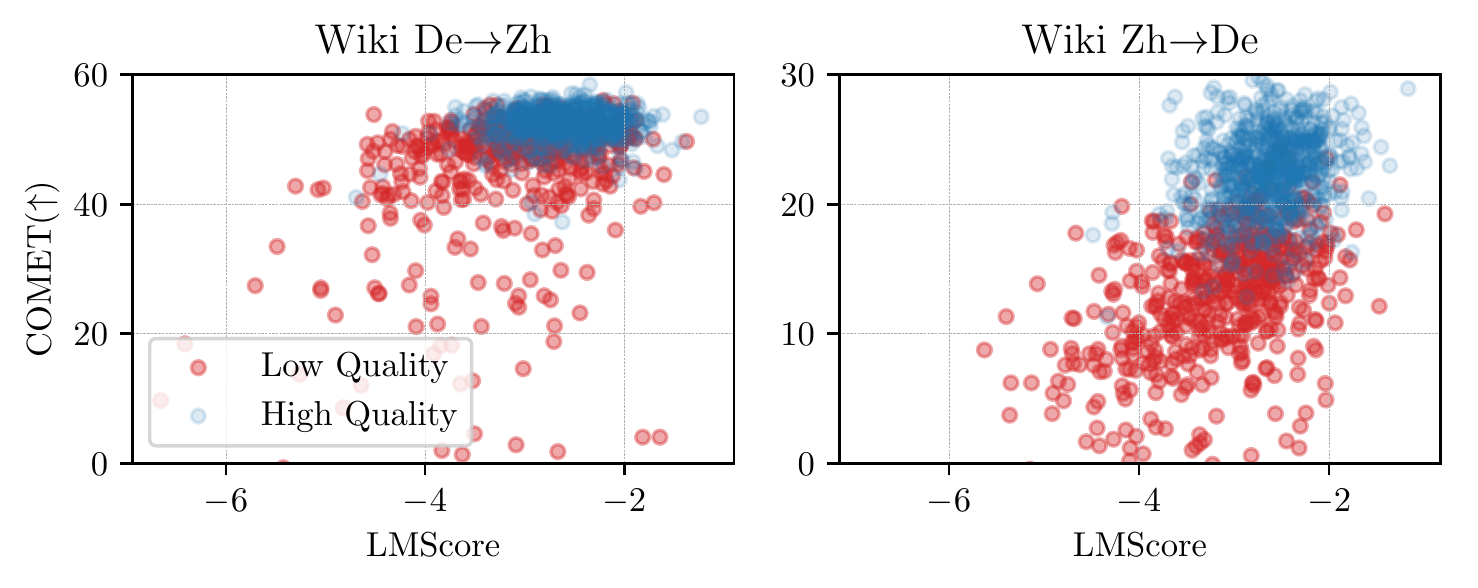}}
    }
    \subcaptionbox{\label{fig:1_shot_factor_correlation_bleu_extra} BLEU vs. LMScore}{
        \resizebox{\columnwidth}{!}{\includegraphics[scale=0.50]{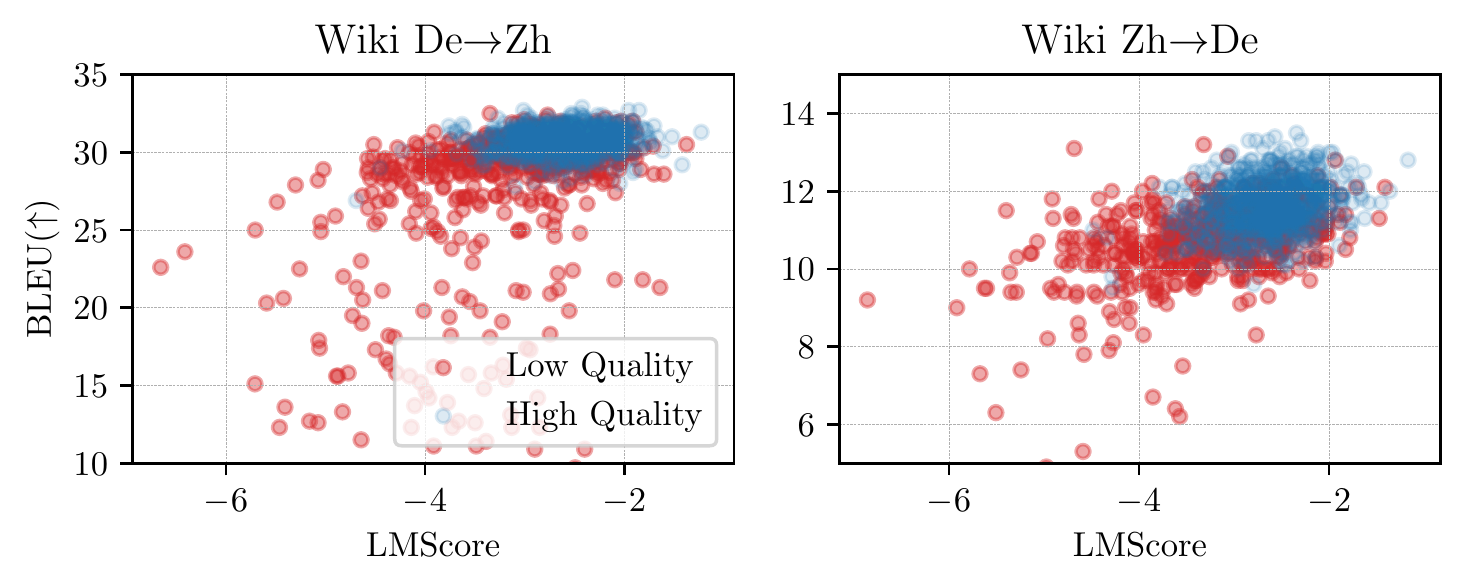}}
    }
    \caption{\label{fig:1_shot_factor_correlation_extra} Scatter plotting between COMET/BLEU and LMScore for \textit{1-shot} prompting on Wiki De$\leftrightarrow$Zh Ablation sets.}
\end{figure*}

\begin{table*}[t]
    \centering
    \small
    \begin{tabular}{llp{11cm}}
    \toprule
     \multirow{5}{*}{En$\rightarrow$Zh}
     & \multirow{1}{*}{Source} & \chinese{Coordinates: 19°43′10″S 63°18′00″E / 19.71944°S 63.30000°E / -19.71944; 63.30000} \\
     & \multirow{1}{*}{Target} & \chinese{坐标：19°43′10″S 63°18′00″E / 19.71944°S 63.30000°E / -19.71944; 63.30000} \\
     \cmidrule(lr){2-3}
     & Source & SAO 40012 is HD 277559. \\
     & Target & \chinese{SAO 40012是HD 277559。} \\
     \midrule
     \multirow{4}{*}{En$\rightarrow$De}
     & Source & 2002 and 2004. \\
     & Target & 2002 und 2004. \\
     \cmidrule(lr){2-3}
     & Source & Brinton, Lauren and Leslie Arnovick. \\
     & Target & Brinton, Lauren und Leslie Arnovick. \\
    \bottomrule
    \end{tabular}
    \caption{\label{tab:sem_uninfor} Top-ranked parallel examples according to SemScore on WikiMatrix.v1 En-De and En-Zh. Despite showing high semantic similarity, these examples are not very informative. We thus dropped them at selection.}
\end{table*}

\begin{table*}[t]
    \centering
    \small
    \setlength{\tabcolsep}{3.8pt}
    \begin{tabular}{lrrrrrrrrrrrrrr}
    \toprule
    \multirow{3}{*}{Method} & \multicolumn{7}{c}{BLEU} & \multicolumn{7}{c}{COMET} \\
    \cmidrule(lr){2-8} \cmidrule(lr){9-15} 
    & \multicolumn{2}{c}{De $\leftrightarrow$ En} & \multicolumn{2}{c}{De $\leftrightarrow$ Zh} & \multicolumn{2}{c}{En $\leftrightarrow$ Zh} & \multirow{2}{*}{Avg} & \multicolumn{2}{c}{De $\leftrightarrow$ En} & \multicolumn{2}{c}{De $\leftrightarrow$ Zh} & \multicolumn{2}{c}{En $\leftrightarrow$ Zh} & \multirow{2}{*}{Avg} \\
    \cmidrule(lr){2-3} \cmidrule(lr){4-5} \cmidrule(lr){6-7}
    \cmidrule(lr){9-10} \cmidrule(lr){11-12} \cmidrule(lr){13-14}
    & \multicolumn{1}{c}{$\rightarrow$} & \multicolumn{1}{c}{$\leftarrow$} & \multicolumn{1}{c}{$\rightarrow$} & \multicolumn{1}{c}{$\leftarrow$} & \multicolumn{1}{c}{$\rightarrow$} & \multicolumn{1}{c}{$\leftarrow$} & & \multicolumn{1}{c}{$\rightarrow$} & \multicolumn{1}{c}{$\leftarrow$} & \multicolumn{1}{c}{$\rightarrow$} & \multicolumn{1}{c}{$\leftarrow$} & \multicolumn{1}{c}{$\rightarrow$} & \multicolumn{1}{c}{$\leftarrow$} \\
    \midrule
    Zero-Shot & 37.80 & 20.50 & 21.70 & 9.60 & 28.60 & 26.30 & 24.08 & 68.30 & 29.96 & 2.80 & 10.05 & 29.17 & 63.25 & 33.92 \\
    \midrule
    \multicolumn{10}{l}{\it 1-Shot Translation (high-quality pool)}  \vspace{0.1cm} \\
    Random & 37.67\hidden{0.25} & 21.23\hidden{0.25} & 28.70\hidden{0.36} & 9.07\hidden{0.17} & 34.87\hidden{1.07} & 26.30\hidden{0.08} & 26.31\hidden{0.36} & 68.77\hidden{0.57} & 35.56\hidden{0.52} & 47.23\hidden{0.82} & 11.75\hidden{1.33} & 60.69\hidden{1.89} & 65.75\hidden{0.80} & 48.29\hidden{0.99} \\ 
    SemScore & \underline{38.40}\hidden{0.43} & 21.37\hidden{0.12} & \underline{29.17}\hidden{0.12} & \underline{9.47}\hidden{0.17} & 35.50\hidden{0.29} & 26.50\hidden{0.22} & \underline{26.73}\hidden{0.23} & \underline{69.04}\hidden{0.45} & 36.06\hidden{1.45} & \underline{48.79}\hidden{0.96} & \underline{14.63}\hidden{0.86} & 60.54\hidden{0.48} & \underline{66.98}\hidden{0.54} & \underline{49.34}\hidden{0.79} \\  
    LMScore & 37.80\hidden{0.57} & 21.43\hidden{0.12} & 28.13\hidden{0.82} & 9.40\hidden{0.42} & 35.40\hidden{0.37} & \underline{26.73}\hidden{0.33} & 26.48\hidden{0.44} & 68.55\hidden{0.31} & 35.49\hidden{0.13} & 43.54\hidden{2.12} & 13.14\hidden{0.98} & 59.84\hidden{1.62} & \underline{66.98}\hidden{0.38} & 47.92\hidden{0.92} \\ 
    TLength & 37.00\hidden{1.08} & \underline{21.80}\hidden{0.08} & 28.57\hidden{0.33} & \underline{9.47}\hidden{0.21} & \underline{35.90}\hidden{0.29} & 26.53\hidden{0.31} & 26.54\hidden{0.38} & 67.79\hidden{0.50} & \underline{37.00}\hidden{0.68} & 45.66\hidden{1.86} & 13.63\hidden{1.08} & \underline{61.87}\hidden{1.35} & 66.45\hidden{0.68} & 48.73\hidden{1.02} \\
    \midrule
    \multicolumn{10}{l}{\it 5-Shot Translation (high-quality pool)}  \vspace{0.1cm} \\
    Random & \textbf{39.03}\hidden{0.42} & 22.00\hidden{0.22} & 29.37\hidden{0.21} & 10.07\hidden{0.12} & \textbf{37.07}\hidden{0.34} & \textbf{27.20}\hidden{0.22} & \textbf{27.46}\hidden{0.25} & \textbf{70.30}\hidden{0.25} & 36.46\hidden{0.71} & 51.77\hidden{0.16} & 16.74\hidden{1.13} & 63.77\hidden{0.38} & 67.62\hidden{0.76} & 51.11\hidden{0.57} \\ 
    SemScore & 38.13\hidden{0.42} & 21.93\hidden{0.05} & \textbf{30.50}\hidden{0.14} & \textbf{10.20}\hidden{0.14} & 36.87\hidden{0.19} & 26.50\hidden{0.24} & 27.36\hidden{0.20} & 70.12\hidden{0.07} & \textbf{38.40}\hidden{0.06} & \textbf{52.29}\hidden{0.75} & \textbf{16.88}\hidden{0.59} & \textbf{64.40}\hidden{0.28} & \textbf{67.85}\hidden{0.21} & \textbf{51.66}\hidden{0.32} \\ 
    LMScore & 38.87\hidden{0.26} & \textbf{22.03}\hidden{0.05} & 30.20\hidden{0.14} & 9.97\hidden{0.09} & 35.83\hidden{0.76} & 26.13\hidden{0.17} & 27.17\hidden{0.25} & 69.74\hidden{0.00} & 37.01\hidden{0.65} & 51.01\hidden{0.13} & 16.63\hidden{0.75} & 61.74\hidden{0.83} & 67.74\hidden{0.02} & 50.65\hidden{0.40} \\ 
    TLength & 38.57\hidden{0.12} & 22.00\hidden{0.08} & 29.50\hidden{0.24} & 10.00\hidden{0.00} & 35.90\hidden{0.65} & 26.53\hidden{0.33} & 27.08\hidden{0.24} & 68.94\hidden{0.31} & 37.16\hidden{0.35} & 50.80\hidden{0.37} & 15.80\hidden{0.20} & 63.01\hidden{0.30} & 67.29\hidden{0.16} & 50.50\hidden{0.28} \\ 
    \midrule
    \multicolumn{10}{l}{\it 1-shot Translation (Low-quality Pool)}  \vspace{0.1cm} \\
    Random & 36.73\hidden{0.78} & 20.53\hidden{0.54} & \underline{22.23}\hidden{6.46} & 8.23\hidden{0.60} & \underline{34.63}\hidden{1.47} & 26.13\hidden{0.91} & 24.75\hidden{1.79} & 66.82\hidden{1.16} & \underline{34.15}\hidden{0.32} & \underline{10.11}\hidden{40.30} & -1.94\hidden{12.33} & 57.97\hidden{1.81} & 66.08\hidden{0.98} & 38.86\hidden{9.48} \\ 
    Ours & \underline{37.90}\hidden{0.33} & \underline{21.27}\hidden{0.24} & 20.50\hidden{6.66} & \underline{9.37}\hidden{0.17} & 34.47\hidden{0.95} & \underline{26.17}\hidden{0.12} & \underline{24.94}\hidden{1.41} & \underline{68.46}\hidden{0.72} & 33.78\hidden{0.78} & 0.19\hidden{43.86} & \underline{12.07}\hidden{1.75} & \underline{58.05}\hidden{1.53} & \underline{66.75}\hidden{0.47} & \underline{39.88}\hidden{8.18} \\ 
    \bottomrule
    \end{tabular}
    \caption{\label{tab:detailed_res_wiki} Detailed test results for \textit{zero-shot and few-shot} prompting on Wiki Full sets with different selection strategies. \textit{Ours}: the proposed combined strategy; \textit{Random}: random sampling; \textit{SemScore, LMScore} and \textit{TLength} denote selecting top-ranked examples based on the corresponding feature values. We select 3 demonstrations for each setup and report the average. \textit{Avg}: average result over language pairs. \underline{Underlined} results denote the best in each section, while \textbf{Bold} results are the overall best.}
\end{table*}

\begin{table*}[t]
    \centering
    \small
    \begin{tabular}{lrrrrrrrrrr}
    \toprule
    \multirow{3}{*}{Method} & \multicolumn{5}{c}{BLEU} & \multicolumn{5}{c}{COMET} \\
    \cmidrule(lr){2-6} \cmidrule(lr){7-11} 
    & \multicolumn{2}{c}{De $\leftrightarrow$ En} & \multicolumn{2}{c}{En $\leftrightarrow$ Zh} & \multirow{2}{*}{Avg} & \multicolumn{2}{c}{De $\leftrightarrow$ En} & \multicolumn{2}{c}{En $\leftrightarrow$ Zh} & \multirow{2}{*}{Avg} \\
    \cmidrule(lr){2-3} \cmidrule(lr){4-5}
    \cmidrule(lr){7-8} \cmidrule(lr){9-10}
    & \multicolumn{1}{c}{$\rightarrow$} & \multicolumn{1}{c}{$\leftarrow$} & \multicolumn{1}{c}{$\rightarrow$} & \multicolumn{1}{c}{$\leftarrow$} & & \multicolumn{1}{c}{$\rightarrow$} & \multicolumn{1}{c}{$\leftarrow$} & \multicolumn{1}{c}{$\rightarrow$} & \multicolumn{1}{c}{$\leftarrow$} \\
    \midrule
    Zero-Shot & \textbf{28.30} & 15.70 & 20.70 & 16.80 & 20.38 & 46.01 & 13.32 & 4.63 & 7.92 & 17.97 \\
    \midrule
    \multicolumn{10}{l}{\it 1-Shot Translation (high-quality pool)}  \vspace{0.1cm} \\
    Random & 25.63\hidden{0.59} & \underline{16.37}\hidden{0.46} & 26.03\hidden{0.12} & 17.03\hidden{0.95} & 21.27\hidden{0.53} & 45.90\hidden{0.93} & 16.89\hidden{0.68} & 40.88\hidden{3.37} & 19.14\hidden{5.89} & 30.70\hidden{2.72} \\
    SemScore & 26.90\hidden{0.92} & 16.03\hidden{0.09} & \underline{26.30}\hidden{0.43} & \underline{18.07}\hidden{0.70} & \underline{21.82}\hidden{0.54} & \underline{46.39}\hidden{1.94} & 15.13\hidden{1.54} & 41.13\hidden{1.06} & \underline{22.49}\hidden{2.56} & \underline{31.28}\hidden{1.78} \\
    LMScore & \underline{27.53}\hidden{0.34} & 15.70\hidden{0.22} & 25.43\hidden{0.83} & 17.70\hidden{-1.00} & 21.59\hidden{0.10} & 47.47\hidden{0.05} & 17.53\hidden{0.46} & 38.95\hidden{1.85} & 19.29\hidden{-1.00} & 30.81\hidden{0.34} \\
    TLength & 25.60\hidden{0.16} & 16.33\hidden{0.05} & 25.80\hidden{0.33} & 17.43\hidden{1.00} & 21.29\hidden{0.38} & 43.47\hidden{2.06} & \underline{18.24}\hidden{0.34} & \underline{42.17}\hidden{0.58} & 18.82\hidden{2.80} & 30.68\hidden{1.44} \\
    \midrule
    \multicolumn{10}{l}{\it 5-Shot Translation (high-quality pool)}  \vspace{0.1cm} \\
    Random & 26.40\hidden{0.62} & \textbf{17.10}\hidden{0.08} & 26.23\hidden{0.29} & 17.53\hidden{0.94} & 21.82\hidden{0.48} & 48.36\hidden{1.23} & 20.19\hidden{0.13} & 43.97\hidden{0.46} & 22.95\hidden{3.21} & 33.87\hidden{1.26} \\
    SemScore & \underline{27.30}\hidden{0.42} & 16.57\hidden{0.17} & \textbf{26.93}\hidden{0.09} & 18.67\hidden{0.29} & \textbf{22.37}\hidden{0.24} & \textbf{49.33}\hidden{1.02} & 18.83\hidden{0.60} & 43.49\hidden{0.34} & 25.54\hidden{0.94} & 34.30\hidden{0.72} \\
    LMScore & 25.90\hidden{0.08} & 16.87\hidden{0.21} & 26.47\hidden{0.25} & \underline{18.93}\hidden{0.45} & 22.04\hidden{0.25} & 47.77\hidden{0.52} & \textbf{20.83}\hidden{0.20} & 44.76\hidden{0.31} & \textbf{27.41}\hidden{1.12} & \textbf{35.19}\hidden{0.53} \\
    TLength & 25.80\hidden{0.73} & 17.03\hidden{0.12} & 26.55\hidden{-1.00} & 17.63\hidden{0.46} & 21.75\hidden{0.08} & 47.34\hidden{0.71} & 20.78\hidden{0.92} & \textbf{45.17}\hidden{-1.00} & 23.85\hidden{1.62} & 34.29\hidden{0.56} \\
    \midrule
    \multicolumn{10}{l}{\it 1-shot Translation (Low-quality Pool)}  \vspace{0.1cm} \\
    Random & 27.33\hidden{1.72} & 15.53\hidden{0.25} & \underline{25.30}\hidden{1.04} & 20.07\hidden{0.42} & 22.06\hidden{0.86} & 45.29\hidden{3.76} & 14.21\hidden{1.26} & \underline{36.83}\hidden{1.55} & 26.49\hidden{2.42} & 30.70\hidden{2.25} \\ 
    Ours & \underline{27.63}\hidden{0.60} & \underline{15.97}\hidden{0.12} & 25.23\hidden{0.61} & \textbf{20.10}\hidden{0.36} & \underline{22.23}\hidden{0.42} & \underline{47.16}\hidden{0.77} & \underline{15.01}\hidden{1.38} & 34.48\hidden{3.77} & \underline{26.82}\hidden{0.55} & \underline{30.87}\hidden{1.62} \\ 
    \bottomrule
    \end{tabular}
    \caption{\label{tab:detailed_res_wmt} Detailed test results on WMT Full sets.}
\end{table*}

\begin{figure*}[t]
    \centering
    \subcaptionbox{\label{fig:mono_comet_extra} COMET}{
        \resizebox{\columnwidth}{!}{\includegraphics[scale=0.50]{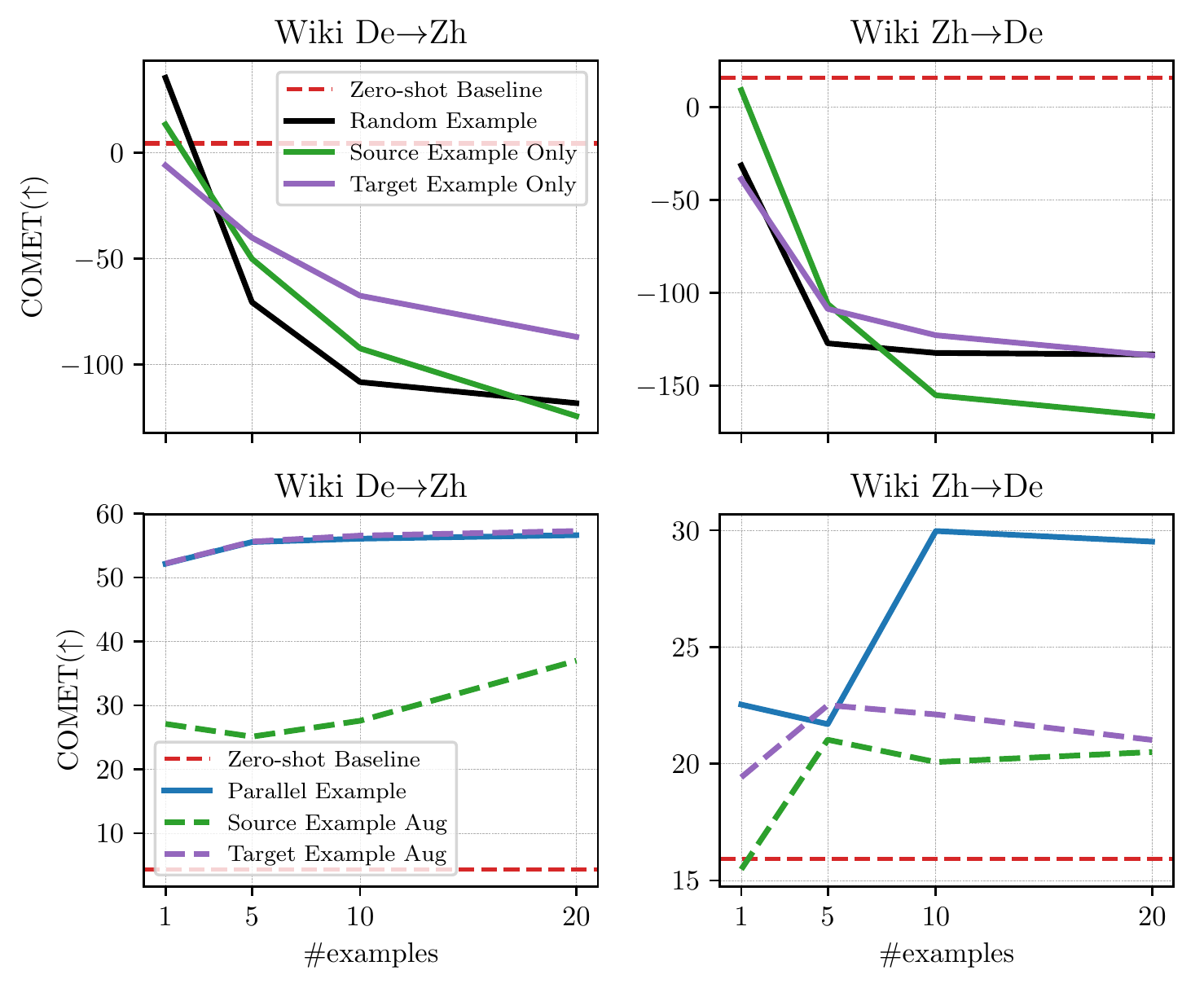}}
    }
    \subcaptionbox{\label{fig:mono_bleu_extra} BLEU}{
        \resizebox{\columnwidth}{!}{\includegraphics[scale=0.50]{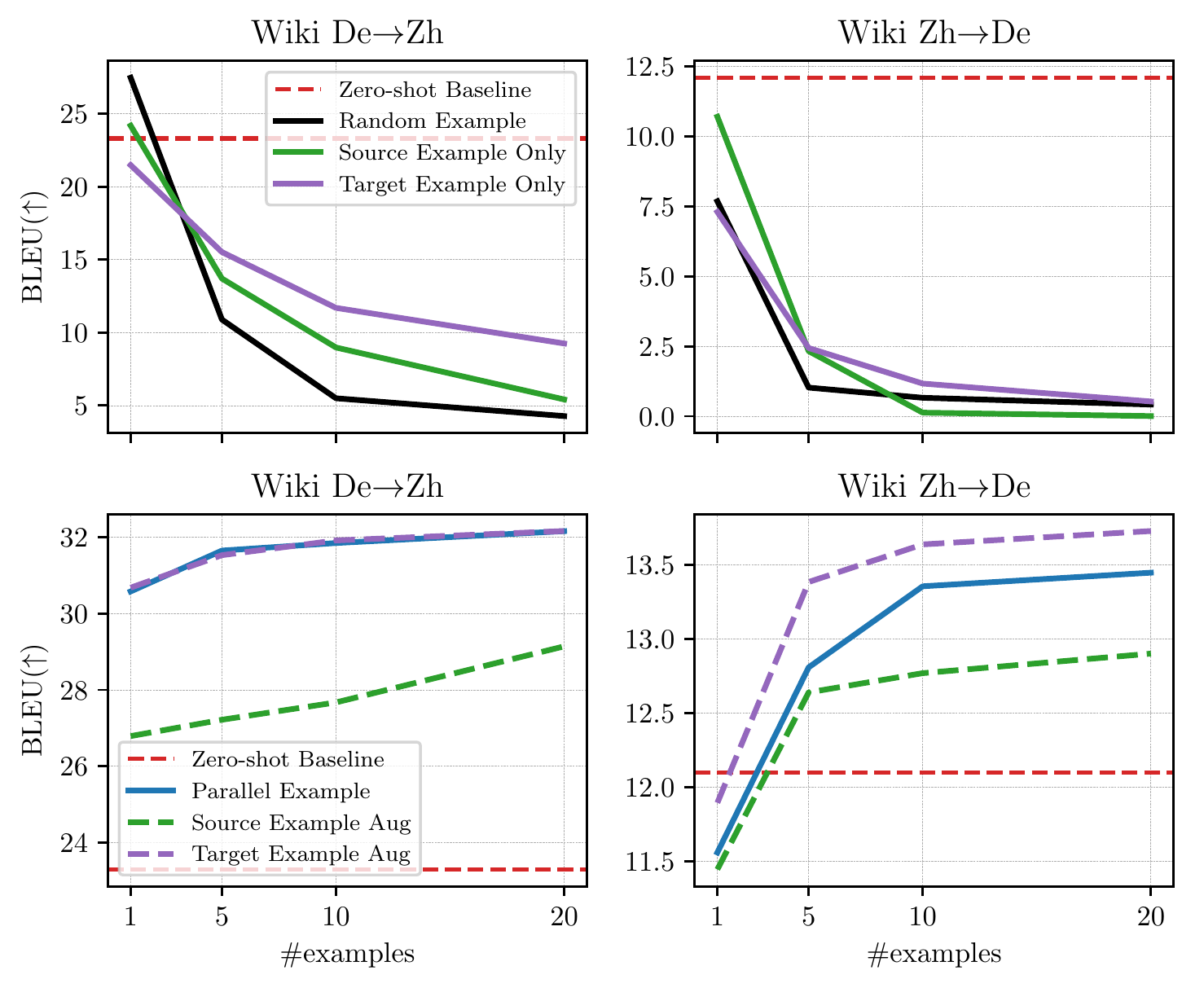}}
    }
    \caption{\label{fig:mono_extra} Results for \textit{few-shot} prompting with monolingual data on Wiki Ablation sets for De$\leftrightarrow$Zh. }
\end{figure*}

\begin{figure*}[t]
  \centering
  \small
  \resizebox{\textwidth}{!}{\includegraphics[scale=0.50]{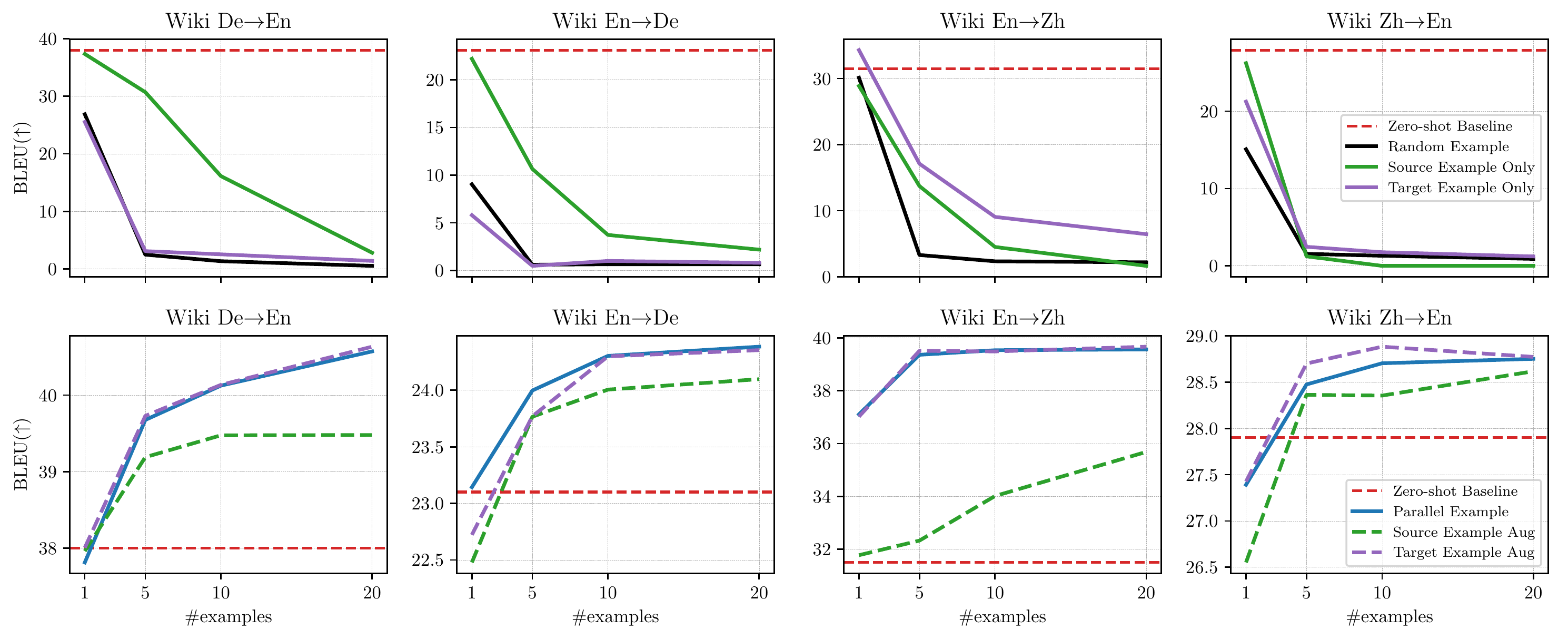}}
  \caption{\label{fig:mono_bleu} BLEU scores for \textit{few-shot} prompting with monolingual data on Wiki Ablation sets.}
\end{figure*}

\begin{table*}[t]
    \centering
    \small
    \setlength{\tabcolsep}{3.8pt}
    \begin{tabular}{llrrrrrrrrrrrr}
    \toprule
    \multicolumn{2}{c}{\multirow{3}{*}{Method}} & \multicolumn{6}{c}{BLEU} & \multicolumn{6}{c}{COMET} \\
    \cmidrule(lr){3-8} \cmidrule(lr){9-14} 
    & & \multicolumn{2}{c}{De $\leftrightarrow$ En} & \multicolumn{2}{c}{De $\leftrightarrow$ Zh} & \multicolumn{2}{c}{En $\leftrightarrow$ Zh} & \multicolumn{2}{c}{De $\leftrightarrow$ En} & \multicolumn{2}{c}{De $\leftrightarrow$ Zh} & \multicolumn{2}{c}{En $\leftrightarrow$ Zh}  \\
    \cmidrule(lr){3-4} \cmidrule(lr){5-6} \cmidrule(lr){7-8}
    \cmidrule(lr){9-10} \cmidrule(lr){11-12} \cmidrule(lr){13-14}
    & & \multicolumn{1}{c}{$\rightarrow$} & \multicolumn{1}{c}{$\leftarrow$} & \multicolumn{1}{c}{$\rightarrow$} & \multicolumn{1}{c}{$\leftarrow$} & \multicolumn{1}{c}{$\rightarrow$} & \multicolumn{1}{c}{$\leftarrow$} & \multicolumn{1}{c}{$\rightarrow$} & \multicolumn{1}{c}{$\leftarrow$} & \multicolumn{1}{c}{$\rightarrow$} & \multicolumn{1}{c}{$\leftarrow$} & \multicolumn{1}{c}{$\rightarrow$} & \multicolumn{1}{c}{$\leftarrow$} \\
    \midrule
    \parbox[t]{2mm}{\multirow{6}{*}{\rotatebox[origin=c]{90}{Prompt Language}}} & 
    De$\rightarrow$En	 & - 	 & \nosig{ 0.06}	 & \nosig{ 0.08}	 & \lessthree{ 0.12}$^\dagger$	 & \lessthree{ 0.13}$^\dagger$	 & \lessthree{ 0.13}$^\dagger$	& -	 & \nosig{-0.02}	 & \nosig{ 0.09}	 & \lessthree{ 0.12}$^\dagger$	 & \nosig{-0.01}	 & \lessthree{ 0.21}$^\ddagger$	\\
    & En$\rightarrow$De	 & \nosig{ 0.07}	 & -	 & \lessthree{ 0.14}$^\ddagger$	 & \lessthree{ 0.19}$^\ddagger$	 & \lessthree{ 0.17}$^\ddagger$	 & \lessthree{ 0.11}$^\dagger$	& \nosig{ 0.01}	 & -	 & \nosig{ 0.07}	 & \lessthree{ 0.21}$^\ddagger$	 & \lessthree{ 0.14}$^\ddagger$	 & \lessthree{ 0.17}$^\ddagger$	\\
    & De$\rightarrow$Zh	 & \nosig{-0.08}	 & \nosig{ 0.06}	 & -	 & \lessthree{ 0.14}$^\ddagger$	 & \lessthree{ 0.24}$^\ddagger$	 & \nosig{-0.05}	& \nosig{ 0.02}	 & \lessthree{ 0.15}$^\ddagger$	 & -	 & \nosig{ 0.08}	 & { 0.40}$^\ddagger$	 & \nosig{ 0.02}	\\
    & Zh$\rightarrow$De	 & \nosig{ 0.00}	 & \lessthree{ 0.26}$^\ddagger$	 & \lessthree{ 0.26}$^\ddagger$	 & -	 & \nosig{ 0.05}	 & \nosig{ 0.01}	 & \nosig{-0.03}	 & \lessthree{ 0.21}$^\ddagger$	 & \lessthree{ 0.22}$^\ddagger$	 & -	 & \lessthree{ 0.13}$^\dagger$	 & \lessthree{ 0.15}$^\ddagger$	\\
    & En$\rightarrow$Zh	 & \nosig{ 0.01}	 & \nosig{-0.01}	 & \lessthree{ 0.24}$^\ddagger$	 & \lessthree{ 0.25}$^\ddagger$	 & -	 & \lessthree{ 0.19}$^\ddagger$	& \nosig{ 0.04}	 & \nosig{-0.01}	 & \lessthree{ 0.22}$^\ddagger$	 & \lessthree{ 0.21}$^\ddagger$	 & -	 & \nosig{ 0.03} \\
    & Zh$\rightarrow$En	 & \lessthree{ 0.15}$^\ddagger$	 & \lessthree{-0.16}$^\ddagger$	 & \lessthree{ 0.14}$^\ddagger$	 & { 0.34}$^\ddagger$	 & \lessthree{ 0.15}$^\ddagger$	 & -  & \lessthree{ 0.25}$^\ddagger$	 & \nosig{ 0.09}	 & \lessthree{ 0.14}$^\ddagger$	 & \lessthree{ 0.21}$^\ddagger$	 & \nosig{ 0.03}	 & -	\\
    \bottomrule
    \end{tabular}
    \caption{\label{tab:res_cross_lingual_detailed_corr} Detailed Spearman's $\rho$ for cross-lingual transfer under \textit{1-shot} prompting on Wiki Ablation sets.  \nosig{Gray} cells indicate insignificance.}
\end{table*}

\begin{table*}[t]
    \centering
    \small
    \setlength{\tabcolsep}{3.8pt}
    \begin{tabular}{llrrrrrrrrrrrr}
    \toprule
    \multicolumn{2}{c}{\multirow{3}{*}{Method}} & \multicolumn{6}{c}{BLEU} & \multicolumn{6}{c}{COMET} \\
    \cmidrule(lr){3-8} \cmidrule(lr){9-14} 
    & & \multicolumn{2}{c}{De $\leftrightarrow$ En} & \multicolumn{2}{c}{De $\leftrightarrow$ Zh} & \multicolumn{2}{c}{En $\leftrightarrow$ Zh} & \multicolumn{2}{c}{De $\leftrightarrow$ En} & \multicolumn{2}{c}{De $\leftrightarrow$ Zh} & \multicolumn{2}{c}{En $\leftrightarrow$ Zh}  \\
    \cmidrule(lr){3-4} \cmidrule(lr){5-6} \cmidrule(lr){7-8}
    \cmidrule(lr){9-10} \cmidrule(lr){11-12} \cmidrule(lr){13-14}
    & & \multicolumn{1}{c}{$\rightarrow$} & \multicolumn{1}{c}{$\leftarrow$} & \multicolumn{1}{c}{$\rightarrow$} & \multicolumn{1}{c}{$\leftarrow$} & \multicolumn{1}{c}{$\rightarrow$} & \multicolumn{1}{c}{$\leftarrow$} & \multicolumn{1}{c}{$\rightarrow$} & \multicolumn{1}{c}{$\leftarrow$} & \multicolumn{1}{c}{$\rightarrow$} & \multicolumn{1}{c}{$\leftarrow$} & \multicolumn{1}{c}{$\rightarrow$} & \multicolumn{1}{c}{$\leftarrow$} \\
    \midrule
    \parbox[t]{2mm}{\multirow{6}{*}{\rotatebox[origin=c]{90}{Prompt Language}}} 
   &  De$\rightarrow$En	 & - & -0.32 & \posres{ 5.02} & -0.86 & \posres{ 1.29} & \posres{ 0.00} & - & -1.08 & \posres{35.04} & \posres{ 2.71} & \posres{ 7.00} & -0.01 \\
   & En$\rightarrow$De	 & -0.69 & - & \posres{ 3.88} & -0.69 & \posres{ 1.21} & -0.41 & -0.46 & - & \posres{26.01} & \posres{ 1.56} & \posres{ 6.31} & -2.40 \\
   & De$\rightarrow$Zh	 & -0.63 & -0.48 & - & -0.65 & \posres{ 4.38} & \posres{ 0.04} & \posres{ 0.92} & -3.68 & - & \posres{ 4.16} & \posres{23.51} & -0.34 \\
   & Zh$\rightarrow$De	 & -0.66 & -0.86 & \posres{ 6.84} & - & \posres{ 3.23} & \posres{ 0.19} & \posres{ 0.71} & -6.15 & \posres{43.67} & - & \posres{17.54} & \posres{ 0.51} \\
   & En$\rightarrow$Zh	  & -1.54 & -1.17 & \posres{ 6.23} & -1.44 & - & -1.50  & -6.00 & -4.47 & \posres{41.77} & -1.79 & - & -2.20 \\
   & Zh$\rightarrow$En	 & -1.12 & -1.00 & \posres{ 1.78} & -1.11 & \posres{ 4.81} & - & -2.63 & -3.85 & \posres{15.25} & \posres{ 3.90} & \posres{25.29} & - \\
    \bottomrule
    \end{tabular}
    \caption{\label{tab:res_cross_lingual_detailed_quality} Detailed translation results (relative against the zero-shot baseline) for cross-lingual transfer under \textit{1-shot} prompting on Wiki Ablation sets.  \posres{Blue} cells indicate positive gains.}
\end{table*}

\begin{table*}[t]
    \centering
    \small
    \begin{tabular}{llrrr}
    \toprule
    \multicolumn{2}{l}{Transfer from Wiki to $\Rightarrow$} & \multicolumn{1}{c}{WMT} & \multicolumn{1}{c}{IT} & \multicolumn{1}{c}{Medical} \\
    \midrule
    \multirow{2}{*}{Correlation} & En$\rightarrow$De & \nosig{0.05} & \nosig{0.11} & 0.15$^\dagger$ \\
    & De$\rightarrow$En & -0.25$^\ddagger$ & 0.19$^\ddagger$ & \nosig{0.07} \\
    \midrule
    \multirow{2}{*}{$\Delta$ Quality} & En$\rightarrow$De & -0.45 & \posres{+0.88} & -0.21 \\ 
    & De$\rightarrow$En & -0.43 & \posres{+1.00} & \posres{+0.77} \\
    \bottomrule
    \end{tabular}
    \caption{\label{tab:cross_domain_transfer_bleu} Spearman's $\rho$ and relative performance (in BLEU) for cross-domain transfer under \textit{1-shot} prompting.}
\end{table*}

\begin{table*}[t]
    \centering
    \small
    \begin{tabular}{lrrrr}
    \toprule
    \multirow{2}{*}{Setting} & \multicolumn{2}{c}{0-shot} & \multicolumn{2}{c}{1-shot} \\
    \cmidrule(lr){2-3} \cmidrule(lr){4-5}
    & De$\rightarrow$Zh & Zh$\rightarrow$De & De$\rightarrow$Zh & Zh$\rightarrow$De \\
    \midrule
    Direct & 21.70 & 9.60 & 28.70 & 9.07 \\
    Pivoting & \textbf{24.4} & \textbf{11.5} & \textbf{29.47} & \textbf{11.47} \\
    \bottomrule
    \end{tabular}
    \caption{\label{tab:directxy_vs_pivoting_bleu} BLEU scores for direct vs. pivoting translation for De$\leftrightarrow$Zh on Wiki Full sets.}
\end{table*}

\begin{table*}[t]
    \centering
    \small
    \begin{tabular}{lrrrrrrrr}
    \toprule
    \multirow{2}{*}{Method} & \multicolumn{4}{c}{BLEU} & \multicolumn{4}{c}{COMET} \\
    \cmidrule(lr){2-5} \cmidrule(lr){6-9} 
    & IT & Law & Medical & Avg & IT & Law & Medical & Avg \\
    \midrule
    Zero-Shot & 32.4 & 28.5 & 31.3 & 30.7 & 12.39 & 32.85 & 33.99 & 26.41 \\
    \midrule
    \multicolumn{8}{l}{\it 1-shot Translation (Low-quality Pool)}  \vspace{0.1cm} \\
    Random & \underline{33.70}\hidden{0.45} & 27.33\hidden{1.41} & 30.80\hidden{-1.00} & 30.61\hidden{0.29} & 29.12\hidden{3.14} & \underline{30.22}\hidden{8.66} & 34.08\hidden{-1.00} & 31.14\hidden{3.60} \\
    Ours & 32.93\hidden{0.87} & \underline{27.60}\hidden{0.16} & \underline{33.23}\hidden{0.05} & \underline{31.26}\hidden{0.36} & \underline{29.95}\hidden{0.34} & 29.60\hidden{4.50} & \underline{41.37}\hidden{0.08} & \underline{33.64}\hidden{1.64} \\ 
    \midrule
    \multicolumn{8}{l}{\it Cross-domain Transfer}  \vspace{0.1cm} \\
    Wiki$\Rightarrow$Multi-Domain & \underline{32.90}\hidden{0.00} & \underline{26.73}\hidden{0.17} & \underline{31.87}\hidden{0.17} & \underline{30.50}\hidden{0.11} & \underline{25.08}\hidden{1.16} & \underline{33.27}\hidden{0.28} & \underline{37.85}\hidden{0.47} & \underline{32.07}\hidden{0.63} \\ 
    WMT$\Rightarrow$Multi-Domain & 30.87\hidden{2.32} & 25.37\hidden{1.64} & 31.43\hidden{0.42} & 29.22\hidden{1.46} & 12.98\hidden{7.87} & 30.34\hidden{2.74} & 34.80\hidden{1.70} & 26.04\hidden{4.10} \\
    \bottomrule
    \end{tabular}
    \caption{\label{tab:res_multi_domain} Cross-domain transfer results on Multi-Domain Full sets under \textit{1-shot} prompting. We adopt the SemScore-based strategy for example selection using the default Wiki/WMT Full candidate pool. Results are averaged over 3 different demonstrations.}
\end{table*}

\end{document}